\documentclass[10pt,journal,compsoc]{IEEEtran}
\ifCLASSOPTIONcompsoc
  \usepackage[nocompress]{cite}
\else
  \usepackage{cite}
\fi

\usepackage[utf8]{inputenc} 
\usepackage[T1]{fontenc}    
\usepackage{url}            
\usepackage{booktabs}       
\usepackage{amsfonts}       
\usepackage{nicefrac}       
\usepackage{microtype}      
\usepackage[dvipsnames,table,xcdraw]{xcolor}
\usepackage[numbers,sort]{natbib}
\usepackage{multirow}
\usepackage{nicematrix}
\usepackage{graphicx}
\usepackage{bbding}
\usepackage{floatrow}
\usepackage{subcaption}
\newfloatcommand{capbtabbox}{table}[][\FBwidth]
\usepackage[font=small]{caption}

\usepackage{pifont}
\newcommand{\cmark}{\ding{51}}%
\newcommand{\xmark}{\text{\ding{55}}}

\newcommand{\highlight}[1]{\textcolor{black}{\textbf{#1}}}
\usepackage{wrapfig}
\usepackage{hyperref}
\usepackage{ragged2e}

\hypersetup{
    colorlinks=true,
    linkcolor=red,
    citecolor=RoyalBlue,
    filecolor=magenta,      
    urlcolor=cyan,
    }
\ifCLASSINFOpdf
\else
\fi
\begin{document}
%
\title{Contrastive Masked Autoencoders are \\ Stronger Vision Learners}

\author{
Zhicheng Huang \quad Xiaojie Jin \quad Chengze Lu \quad Qibin Hou \\
Ming-Ming Cheng \quad Dongmei Fu \quad Xiaohui Shen \quad Jiashi Feng
\IEEEcompsocitemizethanks{
 \IEEEcompsocthanksitem Z. Huang and D. Fu are with the School of Automation and Electrical Engineering, University of Science and Technology Beijing, Beijing, China. (zhicheng.huang@xs.ustb.edu.cn, fdm\_ustb@ustb.edu.cn)
 \IEEEcompsocthanksitem X. Jin, X. Shen and J. Feng are with Bytedance Inc., China. (\{jinxiaojie, shenxiaohui.kevin, jshfeng\}@bytedance.com)
 \IEEEcompsocthanksitem C. Lu, Q. Hou and M.M. Cheng are with School of Computer Science, Nankai University, Tianjin, China. (czlu919@outlook.com, andrewhoux@gmail.com, cmm@nankai.edu.cn)
 \IEEEcompsocthanksitem Corresponding author: \textbf{Xiaojie Jin} and \textbf{Jiashi Feng}
}

}

\markboth{Journal of \LaTeX\ Class Files,~Vol.~14, No.~8, August~2015}%
{Shell \MakeLowercase{\textit{et al.}}: Bare Demo of IEEEtran.cls for Computer Society Journals}

\IEEEtitleabstractindextext{%
\begin{abstract}
\justifying
Masked image modeling (MIM) has achieved promising results on various vision tasks. However, the limited discriminability of learned representation manifests there is still plenty to go for making a stronger vision learner. Towards this goal, we propose  Contrastive Masked Autoencoders (CMAE), a new self-supervised pre-training method for learning more comprehensive and capable vision representations. By elaboratively unifying contrastive learning (CL) and masked image model (MIM) through novel designs, CMAE leverages their respective advantages and learns representations with both strong instance discriminability and local perceptibility. Specifically, CMAE consists of two branches where the online branch is an asymmetric encoder-decoder and the momentum branch is a momentum updated encoder. During training, the online encoder reconstructs original images from latent representations of masked images to learn holistic features. The momentum encoder, fed with the full images, enhances the feature discriminability via  contrastive learning with its online counterpart. To make CL compatible with MIM, CMAE introduces two new components, i.e. pixel shifting for generating plausible positive views and feature decoder for complementing features of contrastive pairs. Thanks to these novel designs, CMAE effectively improves the representation quality and transfer performance over its MIM counterpart. CMAE achieves the state-of-the-art performance on highly competitive benchmarks of image classification, semantic segmentation and object detection. Notably, CMAE-Base achieves $85.3\%$ top-1 accuracy on ImageNet and $52.5\%$ mIoU on ADE20k, surpassing previous best results by $0.7\%$ and $1.8\%$ respectively. The source code is publicly accessible at \url{https://github.com/ZhichengHuang/CMAE}.

\end{abstract}

}

\maketitle

\IEEEdisplaynontitleabstractindextext

\IEEEpeerreviewmaketitle

\IEEEraisesectionheading{\section{Introduction}\label{sec:introduction}}

%
%
%
%
\IEEEPARstart{M}{asked} image modeling (MIM) \cite{he2022masked,gao2022convmae,xie2022simmim} has been attracting increasing attention recently in the self-supervised learning field, due to its method simplicity and capability of learning rich and holistic representations. Following the idea of masked language modeling in NLP~\cite{devlin2018bert}, they randomly mask a large portion of the training image patches and use an auto-encoder~\cite{hinton1993autoencoders} to reconstruct the original signals (e.g., raw pixels, offline extracted features) of the masked patches. It has been shown in \cite{he2022masked,gao2022convmae,xie2022simmim} that such a simple framework outperforms previous self-supervised learning methods in both ImageNet classification~\cite{deng2009imagenet} and some downstream tasks, like object detection and semantic segmentation.

When we reflect on the success of MIM, it is inevitable to compare it with another well-proven and prevailing SSL method, i.e. contrastive learning (CL)~\cite{oord2018representation, bachman2019learning}. By adopting a simple discriminative idea that pulling closer representations from the same image and pushing away different images, CL methods naturally endow the pretained model with strong instance discriminability. In contrast to CL, MIM focuses more on learning local relations in input image for fulfilling the reconstruction task, instead of modeling the relation among different images~\cite{li2022architecture}. Therefore, it is suspected that MIM is less efficient in learning discriminative representations. This issue has been manifested by experimental results in~\cite{he2022masked,xie2022simmim}. Based on above analysis,  it is thus natural to ask such a question:  \emph{can we leverage contrastive learning to further strengthen the representation learned by MIM methods?} or, in other words, \emph{would MIM methods benefit from contrastive learning?} Along this direction, a few contemporary works attempt to train vision representation models~\cite{zhou2021ibot,tao2022siamese} by simply combining contrastive learning and MIM learning objectives. But they only show marginal performance gain compared to MIM methods.
These results signify that it is non-trivial to fully leverage the advantages of both frameworks. The challenges are ascribed to various distinctions between them, including input augmentations, training objectives, model architectures, etc.

\begin{figure}[tp!]
        \includegraphics[width=0.7\textwidth]{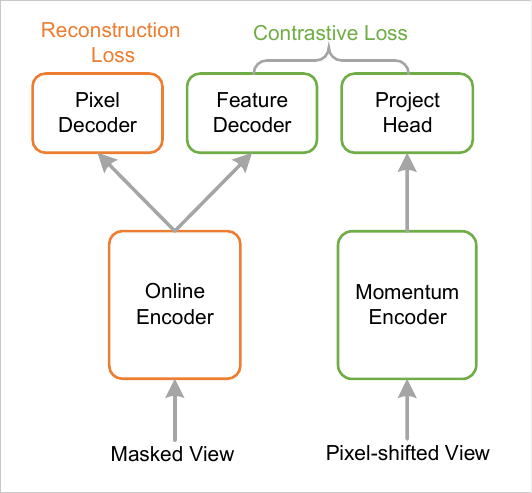}
        \caption{Overview of CMAE. CMAE improves over its MIM counterpart by leveraging contrastive learning through novel designs. To make contrastive learning compatible with MIM, we propose a feature decoder to complement the masked features and a weakly spatial shifting augmentation method for generating plausible contrastive views.}
        \label{fig:overview_teaser}
\end{figure}

To overcome the challenges and learn better image representations for downstream tasks, we aim to explore a possible way to boost the MIM with contrastive learning in a unified framework. With a series of careful studies, we find that input view augmentation and latent feature alignment play important roles in harmonizing MIM and contrastive learning. We thus put dedicated efforts to these components to develop our method. 

An overview of the proposed method is shown in Figure~\ref{fig:overview_teaser}. More specifically, our method introduces a contrastive MAE (CMAE) framework for representation learning. It adopts a siamese architecture~\cite{bromley1993signature}. One branch is an online updated asymmetric encoder-decoder that learns latent representations to reconstruct masked images from a few visible patches, similar to MAE. The other branch is a momentum encoder that provides contrastive learning supervision. To leverage the contrastive learning to improve the feature quality of encoder output, we introduce an auxiliary feature decoder into the online branch, whose output features are used for contrastive learning with the momentum encoder outputs.

We carefully design each CMAE component to enable contrastive learning to benefit the MIM. Different from online encoder whose inputs only contain the visible patches, the CMAE momentum encoder is fed with the full set of image patches. This design ensures semantic integrity of its output features to guide the online encoder. Another notable design choice is: our method uses two decoders, one is to predict the image pixel and perform the MIM task; and another is to recover the features of masked tokens. Since the semantics of each patch are incomplete and ambiguous, it is problematic to use the features of patches directly for contrastive learning. Using an auxiliary feature decoder can address this issue and thus benefit the latent representation learning within the online branch. Moreover, different from existing methods that use strong spatial data augmentations for inputs, we propose a pixel shifting augmentation method for generating more plausible positive views in contrastive learning. Such a simple augmentation is proven effective for improving MIM with contrastive learning. 
\begin{figure}[thp!]
        \includegraphics[width=0.98\textwidth]{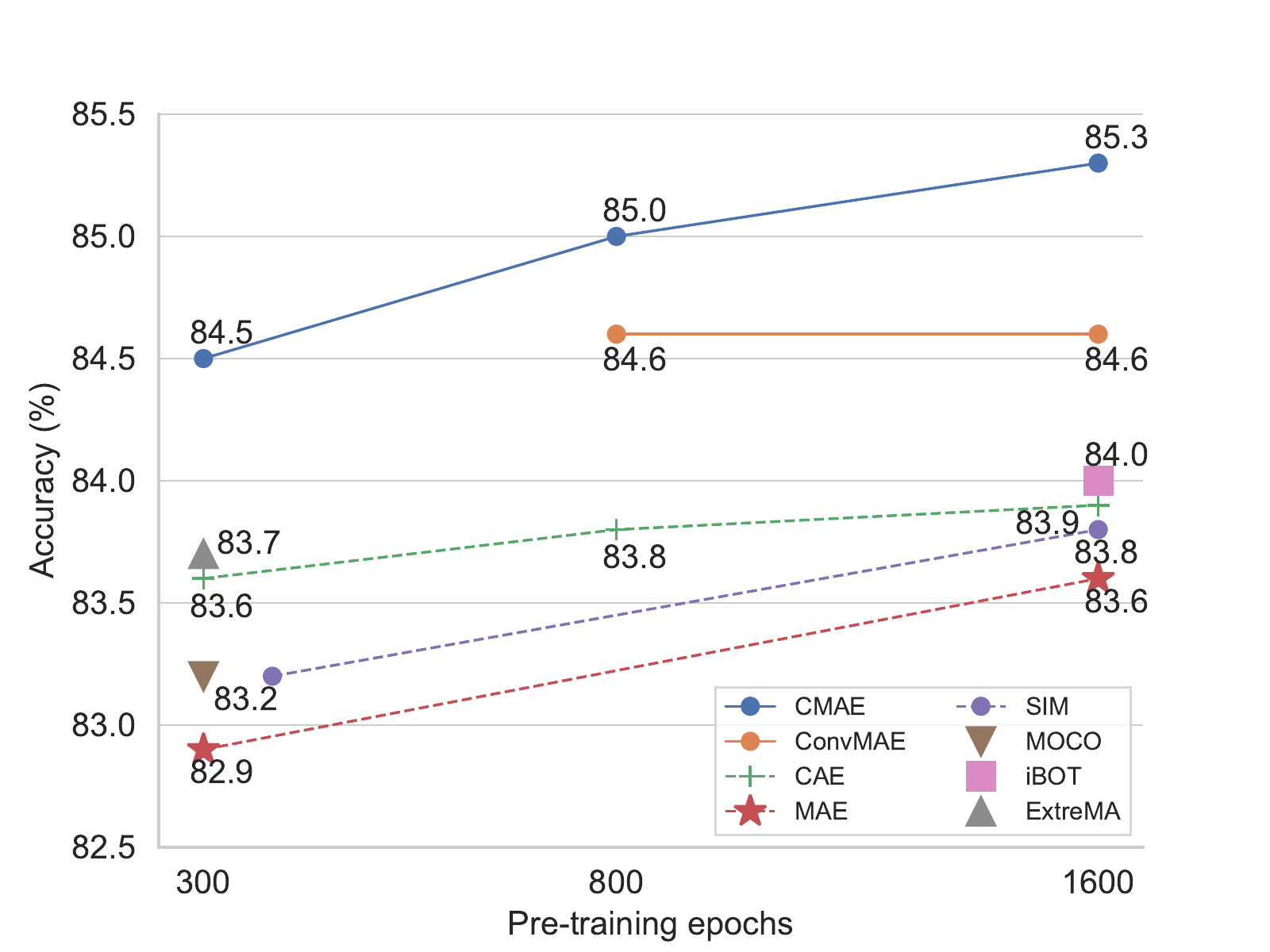}
        \vspace{-4pt}
        \caption{Comparisons with previous state-of-the-art MIM methods on ImageNet-1K in terms of top-1 accuracy at different pre-training epochs.}
        \label{fig:model_pretraining_acc}
\end{figure}
With the above novel designs, the online encoder of our CMAE method can  learn more discriminative features of holistic information and achieve state-of-the-art performance on various pre-training and transfer learning vision tasks. 

Our contributions are summarized as follows. 1) We propose a new CMAE method to explore how to improve the representation of MIM by using contrastive learning. Its learned representations not only preserve the local context sensitive features but also model the instance discriminativeness among different images. 2) To impose contrastive learning upon MIM, we propose a feature decoder to complement the masked features and a weakly spatial shifting augmentation method for generating plausible contrastive views, both of which are effective in improving the encoder feature quality. 3) As shown in Figure \ref{fig:model_pretraining_acc},  our method significantly improves the learned representation of MIM  and sets new state-of-the-art performance. Notably, compared with prior arts, CMAE achieves absolute gains of $0.7\%$ on ImageNet-1k classification validation split, $1.8\%$ mIoU on ADE20K semantic segmentation validation dataset and $0.4\%$ AP$^\text{b}$ and $0.5\%$ AP$^\text{m}$ on CoCo validation split.

\section{Related Work}
Self-supervised learning is attracting increasing attention in computer vision. A bunch of methods have been proposed to advance this technique from different perspectives ~\cite{chen2020simple,grill2020bootstrap,oord2018representation,zbontar2021barlow,bardes2021vicreg,fang2022corrupted,he2022masked,zhou2021ibot}. Broadly speaking, these methods can be categorized into two groups depending on their employed pretext tasks, i.e., contrastive learning~\cite{hadsell2006dimensionality,hjelm2018learning,he2020momentum} and mask image modeling~\cite{he2022masked,xie2022simmim,bao2021beit}.

\begin{figure*}[t]
    \centering
   \includegraphics[width=0.9\textwidth]{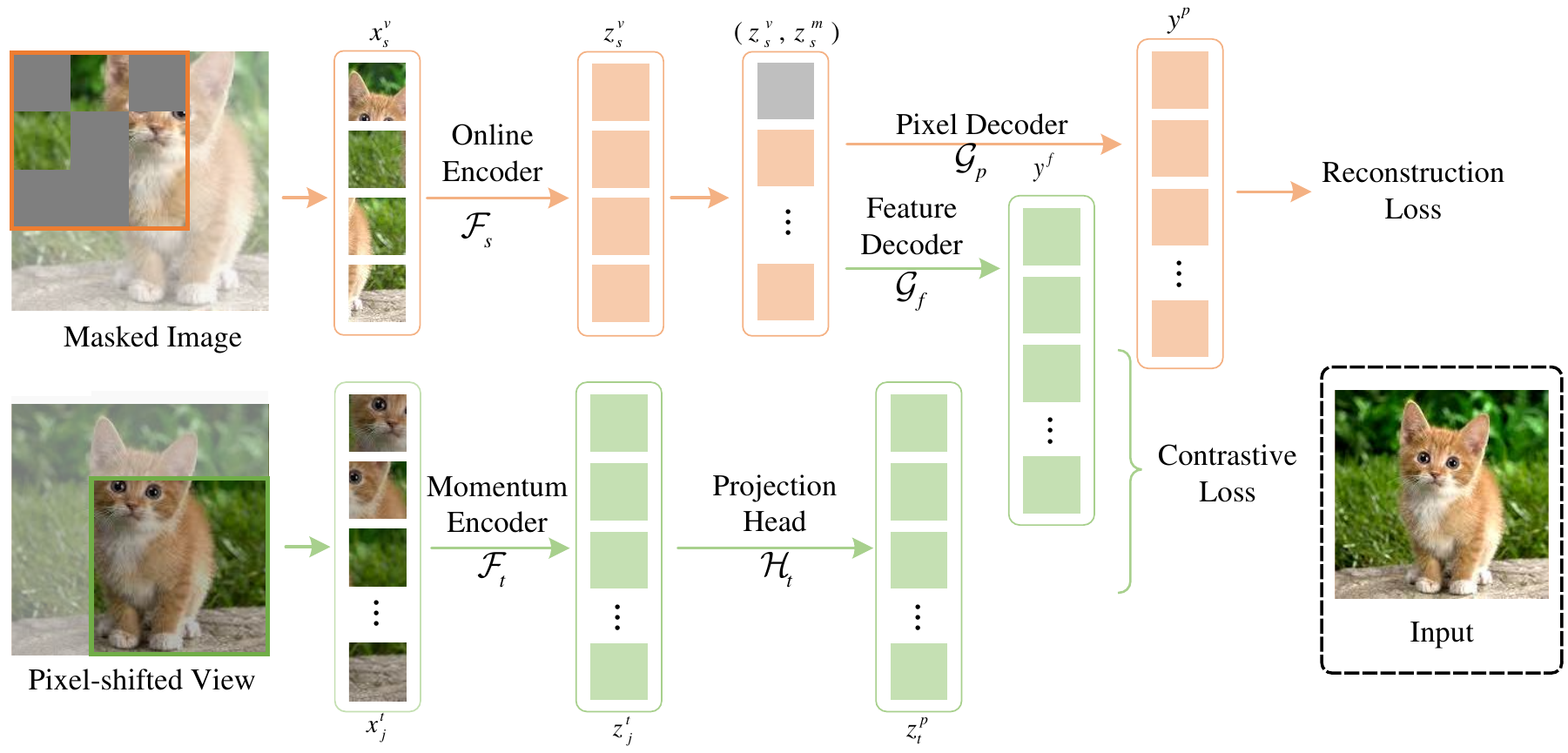}
    \caption{Overall pipeline. Our method contains three components: the online encoder, momentum encoder, and online decoder. Given a training image, it applies \emph{pixel shifting} to generate different views, which are then fed into the online and momentum encoders respectively. The online encoder randomly masks a fraction of the image patches and operates on the visible ones. The momentum encoder operates on the whole view after pixel shifting. The pixel decoder learns to reconstruct the input image from the image tokens (along with MASK tokens) provided by the online encoder, while the \emph{feature decoder} learns to predict the features of the input image for contrastive learning with the momentum encoder output features. During pre-training, the parameters of the momentum encoder and projection head are updated using an exponential moving average algorithm. After the pre-training, only the online encoder is kept for downstream applications. 
    }
    \label{fig:overview}
\end{figure*}

\noindent\textbf{Contrastive learning} aims to learn instance discriminative representations to distinguish an image from the others. This is achieved by pulling together the representation of different views of an individual image and pushing away the other images. Thus most contrastive methods adopt siamese network~\cite{he2020momentum,grill2020bootstrap,chen2021exploring}. To create different view for the same image, a plentiful of methods have been deployed~\cite{oord2018representation,tian2020contrastive,chen2020simple}. Among these methods, data augmentations are most commonly in contrastive learning and investigated in SimCLR.~\cite{chen2020simple}. In practice, contrastive methods rely heavily on a large number of negative samples~\cite{chen2020simple,wu2018unsupervised,tian2020contrastive,he2020momentum}. To better utilize negative samples, MoCo~\cite{he2020momentum} uses a large queue to cache negative examples in memory such that it can take in more negative examples for contrastive learning. BYOL~\cite{grill2020bootstrap} uses an online encoder to predict the output of a momentum encoder, where the momentum encoder is key to avoiding training collapse. To simplify BYOL, SimSiam~\cite{chen2021exploring} proposes the stop-gradient technique to replace the momentum updating. Besides, there are some methods to solve this issue from different views~\cite{caron2018deep,caron2020unsupervised,caron2019unsupervised,zbontar2021barlow}. SwAV~\cite{caron2020unsupervised} proposed an online clustering method and adopted the Sinkhorn-Knopp~\cite{cuturi2013sinkhorn} transform to assist the clustering for each batch. Barlow Twins~\cite{zbontar2021barlow} adopts the cross-correlation matrix to constrain the network to void training collapse. Recently, MoCo-v3~\cite{chen2021empirical} and DINO~\cite{caron2021emerging} are proposed to extend MoCo~\cite{he2020momentum} and BYOL~\cite{grill2020bootstrap} respectively by using Vision Transformer (ViT) as their model backbones. Although contrastive learning methods provide discriminative vision representations, most of them focus on learning global representations while lacking spatial-sensitive representation.

\noindent\textbf{Mask image modeling}~\cite{bao2021beit,chen2020generative,dosovitskiy2020image} is inspired by the success of Masked Language Modeling in NLP~\cite{devlin2018bert,radford2019language} and learns vision representation by constructing the original signal from partial observations. Based on the reconstruction target, these methods can be devided into: pixel-domain reconstruction~\cite{xie2022simmim,he2022masked,wei2022masked,fang2022corrupted,liu2022devil} and auxiliary features/tokens prediction~\cite{bao2021beit,dong2021peco,chen2022context}. 
SimMIM~\cite{xie2022simmim} and MAE~\cite{he2022masked} propose to reconstruct the raw pixel values from either the full set of image patches (SimMIM) or partially observed patches (MAE) to reconstruct the raw image. Compared with SimMIM, MAE is more pre-training efficient because of masking out a large portion of input patches. To learn more semantic features, MaskFeat~\cite{wei2022masked} introduces the low-level local features~(HOG~\cite{dalal2005histograms}) as the reconstruction target, and Ge\textsuperscript{2}-AE adopts the MIM task in frequency domain~\cite{liu2022devil}, while CIM~\cite{fang2022corrupted} opts for more complex input.

Several methods adopt an extra model to generate the target to pre-train the encoder. For instance, BEiT~\cite{bao2021beit} uses the discretized tokens from an offline tokenizer~\cite{ramesh2021zero} to train the encoder. PeCo~\cite{dong2021peco}  instead uses an offline visual vocabulary to guide the encoder. 
Differently, CAE~\cite{chen2022context} uses both the online target and offline network to guide the training of encoder. Furthermore, iBOT~\cite{zhou2021ibot} introduces an online tokenizer to produce the target  to distill the encoder. MVP~\cite{wei2022mvp} adopts a structure identical to BEiT~\cite{bao2021beit} and replaces the tokenizer (e.g. d-VAE in BEiT~\cite{bao2021beit}) with the vision branch of a multimodal model like CLIP that is pre-trained on image-text pairs. Similarly, SIM adopts the siamese network to reconstruct the representations of tokens, based on another masked view~\cite{tao2022siamese}. MSN~\cite{assran2022masked} matches the representation of masked image to that of original image using a set of learnable prototypes. 

Despite MIM models exhibiting favorable optimization properties~\cite{wei2022contrastive} and delivering promising performance, their focus is on learning relationships among the tokens in the input image, rather than modeling the relation among different images as in contrastive learning, which results in less discriminative learned representations. Our method diverges from existing works by proposing innovative designs to fully leverage the advantages of MIM and contrastive learning, thereby providing local context-sensitive representations with the desired discriminativeness for input images.

\section{Method}

\subsection{Framework}
\label{sec:model_structure}

The overall framework of our method is illustrated in Figure~\ref{fig:overview} that consists of three components. The \textit{online encoder and decoder}  learn to reconstruct input images from masked observations. Different from existing MIM methods (e.g.,  MAE~\cite{he2022masked} and SimMIM~\cite{xie2022simmim}), our method further processes the input image via a spatially shifted cropping operation. More importantly, our decoder incorporates an additional feature decoder for predicting the input image features. The \textit{momentum encoder}  transforms the augmented view of the input image into a feature embedding for contrastive learning with the predicted one from the online feature decoder. In this way, the learned representations by the online encoder capture not only holistic features of the input images but also discriminative features, thus achieving better generalization performance. We now elaborate on these components in detail. 

Let us denote the input image $I_s$ to the online encoder as being tokenized into a sequence of $N$ image patch tokens $\{x^{s}_i\}_{i=1}^N$, where $N$ is the total number of image patches. For the masked version of $I_s$, the set of visible tokens is represented as $\{x^v\}$. Similarly, the input image $I_t$ is tokenized into a sequence of image patch tokens $\{x^t_j\}_{j=1}^N$ to serve as the input to the momentum encoder.

\noindent\textbf{Online encoder.}  The online encoder $\mathcal F_s$ maps the visible tokens $x_s^v$ to embedding features $z_{s}^{v}$. Given the token sequence $\{x^{s}_i\}_{i=1}^N$, we mask out a large ratio of patches and feed the visited patches to the online encoder. The online encoder adopts the Vision Transformer (ViT) architecture~\cite{dosovitskiy2020image}, following MAE~\cite{he2022masked}. It first embeds the visible tokens $x_s^v$ by linear projection as token embeddings, and adds the positional embeddings~\cite{vaswani2017attention} $p_s^v$. We feed the fused embedding to a sequence of transformer blocks, and get the embedding features $z_{s}^{v}$.
\begin{equation}
    z_s^v = \mathcal F_s(x_s^v+p_s^v)
\end{equation}
After pre-training, the online encoder $\mathcal F_s$ is used for extracting image representations in downstream tasks.

\noindent\textbf{Momentum encoder.} The momentum encoder is introduced for providing contrastive supervision for the online encoder to learn discriminative representations. Different from existing siamese-based methods~\cite{zhou2021ibot,caron2021emerging}, our momentum encoder $\mathcal F_t$ only serves for contrastive learning, as well as guiding the online encoder to learn more discriminative features. It shares the same architecture as the online encoder~$\mathcal F_s$, but takes the whole image as input, in order to reserve the semantic integrity and the discriminativeness of the learned representations. Using the whole image as input to the momentum encoder is important for the method performance, which is experimentally verified in Section~\ref{ablation}. Unlike tokens in NLP, whose semantic are almost certain, image token is ambiguous in its semantic meaning~\cite{zhou2021ibot}. To avoid ambiguity, we adopt global representations for contrastive learning. The mean-pooled feature of momentum encoder is used for its simplicity, i.e. 
\begin{equation}
    z^t = \frac{1}{N} \sum_{j=1}^N \mathcal F_t(x^t_j),
\end{equation}
where $x^t_j$ is the input token for momentum encoder, $z^t_j$ is the output sequence of momentum encoder and $z^t$ is the feature obtained after performing mean pooling operation on the output sequence $z^t_j$, which is used to represent the input image.

Different from the online encoder, we update parameters of the momentum encoder by exponential moving average (EMA). That is, denoting the parameters of $\mathcal F_s$ and $\mathcal F_t$ as $\theta_s$ and $\theta_t$ respectively, the parameters are updated by $\theta_t \gets \mu \theta_t + (1-\mu )\theta_s$. Here $\mu $ is fixed as 0.996 across the experiments. Momentum update is used since it stabilizes the training by fostering smooth feature changes, as found in MoCo~\cite{he2020momentum} and BYOL~\cite{grill2020bootstrap}.

\noindent\textbf{Online decoder.} The decoder aims to map the latent features $z_s^v$ and MASK token features to the feature space of the momentum encoder and the original images. Specifically, the decoder receives both the encoded visible tokens $z_s^v$ and MASK tokens $z_s^m$. 
 
Similar to MAE~\cite{he2022masked}, position embeddings are added to input tokens. Due to different mapping targets, our online decoder has two branches of decoder structure, one is a pixel decoder, and another is a feature decoder. The pixel decoder $\mathcal G_p$ learns to reconstruct the pixel of the masked patches. We use the full set of tokens, which contains both $z_s^v$ and $z_s^m$, to predict the pixel of patches $y^m$. This module can promote the model to learn holistic representation for each patch in an image. We set the pixel decoder to be stacked transformer blocks:
\begin{equation}
    y^{\prime}_m= \mathbb I \cdot \mathcal G_p(z_s^v, z_s^m),
\end{equation}
where $\mathbb I$ is an indicator to only select the prediction corresponding to masked tokens from output sequence $y^p$, and $y^{\prime}_m$ is the output prediction for the masked patches.

To align with the output of the momentum encoder, feature decoder~$\mathcal G_f$ is applied to recover the feature of masked tokens. The feature decoder has the same structure as the pixel decoder but non-shared parameters for serving a different learning target. The prominence of such design choices will be discussed in the architecture part in Section~\ref{sec:discussion}. Given the encoded visible tokens $z_s^v$, we add the masked tokens $z_s^m$ and use this full set to predict the feature of masked tokens. Similar as done in momentum encoder, we apply the mean pooling operation on the output of feature decoder $y^f$ as the whole image representation~$y_s$, and then use this feature for contrastive learning.
\begin{equation}
    y_s = \frac{1}{N}\sum \mathcal G_f(z_s^v,z_s^m),
\end{equation}
where $N$ is the total number of tokens in the full set.

\begin{table*}[]
    \centering
    \small
    \setlength{\tabcolsep}{7pt}
    \caption{Comparison of CMAE with previous methods on training objective, input generation and architecture. The top-1 accuracy on ImageNet is also presented. Please refer to Section~\ref{sec:discussion} for more detailed explanations.}

    \begin{tabular}{l|ccc|c|cc|c} \toprule
    & \multicolumn{3}{c|}{Training objective}                             & Input                 & \multicolumn{2}{c|}{Architecture}            & \multirow{3}{*}{Accuracy} \\ \cmidrule(lr){2-4} \cmidrule(lr){5-5} \cmidrule(lr){6-7}
    & reconstruct              & intra-view          & inter-image            & positive view                  & feature              & separate             &                     \\
    & loss                 & match                 & contrast              & alignment             & complement           & encoder/decoder       &                     \\ \midrule
MSN~\cite{assran2022masked}     & \xmark & \checkmark  &  \xmark &  \xmark  &  \xmark  &  \xmark & 83.4  \\
ExtreMA~\cite{wu2022extreme}    & \xmark & \checkmark  & \xmark &  \checkmark    &  \xmark  &  \xmark& 83.7  \\
MAE~\cite{he2022masked}         & \checkmark  & \xmark&  \xmark&  N.A.          &  N.A.          &  \checkmark  & 83.6  \\
CAE~\cite{chen2022context}      & \checkmark  & \xmark&  \xmark&  \xmark  &  \xmark  &  \checkmark  & 83.9  \\
iBot~\cite{zhou2021ibot}        & \checkmark  & \checkmark  &  \xmark  &  \xmark  &  \xmark  &  \xmark& 84.0  \\
SIM~\cite{tao2022siamese}       & \checkmark  & \checkmark  &  \checkmark  &  \xmark  &  \xmark  &  \checkmark  & 83.8  \\
\midrule
CMAE    & \checkmark  & \checkmark  &  \checkmark  &  \checkmark    &  \checkmark    &  \checkmark  & 84.7  \\
\bottomrule
\end{tabular}
\label{tab:methods_comp}
\end{table*}

\subsection{View augmentations}
\label{sec:view_aug}

Typically, masking image modeling pre-training tasks only utilizes a single view of the input image, which only contains visited patches. But contrastive learning often adopts two different augmented views. To make MIM and contrastive learning be compatible with each other, our method also generates two different views and feeds them to its online and momentum branches, respectively. 

In contrastive learning, the most commonly used view augmentations can be divided into two types: spatial transfer~(e.g., random resized cropping, flipping) and color transfer~(e.g., color jittering and random grayscaling). For MIM tasks, color enhancements degrade the results~\cite{he2022masked}, so we do not apply them to the input of the online branch. Spatial and color data augmentations are applied to the momentum branch input to avoid a trivial solution. 

We first consider two branches using two different random crops, following the common practice in contrastive learning. However, we observe this recipe has an adverse effect on model performance (refer to Section~\ref{ablation}). We conjecture that this issue is related with the large disparity between the inputs of online/momentum encoders when randomly cropped regions are far apart or scarcely semantic-relevant. Different with using intact paired views in usual contrastive methods, the operation of masking out a large portion of input in MIM may amplify such disparity and therefore creates false positive views. Consequently, performing contrastive learning on these misaligned positive pairs actually incurs noise and hampers the learning of discriminative and meaningful representations. 

To address the above issue, we propose a weakly augmentation method named \textbf{pixel shifting} for generating the inputs of online/momentum encoders. The core idea is first to obtain a master image by a resized random cropping from the original image. Then two branches share the same master image and generate respective views by slightly shifting cropping locations over the master image. In more details, we denote the master image as $I$. The shape of $I$ is $(w+p,h+p,3)$, where $w, h$ is the width and height of target input size for our model and $p$ is the longest shifting range allowed. For online branch, we use the region of $[0:w,0:h,:]$ as our input image~$I_s$. For momentum branch, we use the region of~$[r_w:r_w+w,r_h:r_h+h,:]$ as our input image~$I_t$. $r_w$ and $r_h$ are independent random values in the range of $[0,p)$. Subsequently, we only apply masking operation without color augmentation for the input of online encoder~$I_t$, which is consistent with MAE. For the momentum encoder's input image, we utilize color augmentation but do not apply masking operation. The distinct augmentations for each encoder input generate different views of the same image to facilitate contrastive learning between the online and momentum encoders.

\subsection{Training Objective}
\label{loss_function}
\noindent\textbf{Reconstruction loss.} Following~\cite{he2022masked}, we use the normalized pixel as target in the reconstruction task. We adopt the Mean Squared Error (MSE) as loss function and compute the loss only on masked patches between the pixel decoder prediction and the original image. The math formulation is
\begin{equation}
    L_r = \frac{1}{N_m}\sum(y^{\prime}_m - y_m)^2 ,
\end{equation}
where $N_m$ is the number of masked patches in an image, and $L_r$ is the loss value.

\noindent\textbf{Contrastive loss.} For clarity, we describe the contrastive loss design of our method from two aspects: loss function and head structure. Two widely used loss functions are taken into consideration, i.e. InfoNCE~\cite{he2020momentum,chen2020simple} loss and BYOL-style~\cite{grill2020bootstrap,caron2021emerging} loss. The former seeks to simultaneously pull close positive views from the same sample and push away negative samples while the latter only maximizes the similarity between positive views. Although some recent works find they may be inherently unified~\cite{tao2022exploring}, we still analyze them separately due to their diverse effects on representation learning. In our method, we observe better performance using InfoNCE~\cite{oord2018representation} so we use it defaultly. Details are referred to Section~\ref{ablation}. For the head structure,  we adopt the widely used "projection-prediction" structure following~\cite{chen2021empirical,grill2020bootstrap}. Specifically, we append the "projection-prediction" and "projection" head to feature decoder and momentum encoder respectively. The projection head~$\mathcal H_t$ with momentum encoder is also updated by exponential moving average.
Due to the large differences on generating inputs for online/momentum encoder (refer to Section~\ref{sec:view_aug}), we use asymmetric contrastive loss, which is distinguished from previous methods~\cite{chen2021empirical,grill2020bootstrap}.
The representation from feature decoder $y_s$ is transformed by the "projection-prediction" structure to get $y_{s}^p$. Similarly for the representation from momentum encoder $z^t$, we apply the projection head and get $z_{t}^p$. We then compute the cosine similarity $\rho$ between them: 
\begin{equation}
    \rho=\frac{y_{s}^p\cdot z_{t}^p}{\left \|y_{s}^p \right \|_{2} \left\|z_{t}^p \right \|_{2}}.
\end{equation}
We denote $\rho^+$ as the positive pairs cosine similarity, which is constructed by $y_{s}^p$ and $z_{t}^p$ from the same image. $\rho_{j}^-$ indicates the cosine similarity for the $j$-th negative pair. We use the $z_{t}^p$ from different images in a batch to construct negative pairs. The loss function of InfoNCE loss is
\begin{equation}
    L_c = -\log \frac{\exp(\rho^{+}/\tau)}{\exp(\rho^{+}/\tau)+\sum_{j=1}^{K-1}(\exp(\rho_j^{-} /\tau))},
\label{eq:infonce}
\end{equation}
where $\tau$ is the temperature constant, which is set to $0.07$. $K$ is the batch size.

The overall learning target is a weighted combination of reconstruction loss $L_r$ and contrastive loss $L_c$ defined as:
\begin{equation}
    L = L_r + \lambda_c L_c.
\label{eq:total_loss}
\end{equation}

\begin{table*}[htp!]
    \centering
    \small
    \caption{Comparison of our model with existing methods on ViT-B. We evaluate them with the top-1 accuracy on ImageNet. The symbol of \textsuperscript{*} throughout experiments denotes using convolutions instead of linear transformation as the tokenizer for visual patches. }
    \begin{tabular}{lcccccc}
    \toprule
         Method & Pre-training epochs    & Params.(M)  & Supervision & Accuracy \\
    \midrule
        MoCo-v3~\cite{chen2021empirical} & 300  & 86 & RGB &  83.2  \\
        DINO~\cite{caron2021emerging}    & 300 & 86 & RGB & 82.8 \\
        CIM~\cite{fang2022corrupted}     & 300  & 86 & RGB & 83.3 \\
        BEiT~\cite{bao2021beit}          & 800  & 86 & DALLE & 83.2 \\
        SimMIM~\cite{xie2022simmim}      & 800  & 86 & RGB   & 83.8 \\
        PeCo~\cite{dong2021peco}         & 800  & 86 & Perceptual Codebook & 84.5 \\
        MaskFeat~\cite{wei2022masked}    & 1600 & 86 & HOG & 84.0  \\
        CAE~\cite{chen2022context}       & 1600 & 86 & DALLE+RGB  & 83.9  \\
        iBOT~\cite{zhou2021ibot}         & 1600  & 86 & RGB & 84.0 \\
        SIM~\cite{tao2022siamese}        & 1600 & 86 & RGB & 83.8 \\
        MAE~\cite{he2022masked}         & 1600 & 86 & RGB  & 83.6 \\
        
        CMAE (ours) & 800 & 86 & RGB & 84.4  \\
        CMAE (ours) & 1600 & 86 & RGB & \highlight{84.7} \\
    \midrule
    ConvMAE\textsuperscript{*}~\cite{gao2022convmae} & 800 & 86 & RGB & 84.6 \\
    ConvMAE\textsuperscript{*}~\cite{gao2022convmae} & 1600 & 86 & RGB & 84.6 \\
    CMAE\textsuperscript{*} (ours)  & 800 & 86 & RGB & 85.0 \\
    CMAE\textsuperscript{*} (ours) & 1600 & 86 & RGB & \highlight{85.3} \\
    \bottomrule
        
    \end{tabular}
    
    \label{tab:ImageNet-1k}
\end{table*}
\subsection{Connections and analysis}
\label{sec:discussion}

To elucidate the correlations and distinctions between the CMAE and preceding methodologies, we undertake comparative assessments from various perspectives such as training objective, input, and architecture. The outcomes are demonstrated in Table~\ref{tab:methods_comp}. Our primary focus is on methodologies that utilize either contrastive information in MIM or masked image input. Approaches that solely employ MIM or contrastive learning are not within the purview of this discussion, as they are evidently distinct from our method.

\noindent\textbf{Training objective.} The CMAE leverages both the reconstruction loss and contrastive loss during optimization. As inferred from Eq.~\eqref{eq:infonce}, the contrastive loss in CMAE encompasses both intra-view matching and inter-image contrast. Consequently, the generated representations are encouraged to exhibit desirable characteristics of instance discrimination and spatial sensitivity. In contrast, methodologies such as MSN~\cite{assran2022masked} and ExtreMA~\cite{wu2022extreme}, which have divergent motivations from ours, disregard reconstruction loss and employ masked input for regularization or data augmentation purposes. iBot~\cite{zhou2021ibot} exclusively adopts a distillation loss between positive views by maximizing intra-view matching scores, overlooking contrastive learning with negative samples. Additionally, CMAE only adopts an asymmetric loss for contrastive learning, which is less computationally expensive than iBot. Although SIM~\cite{tao2022siamese} also utilizes both losses, it differs from CMAE in terms of the reconstruction target. While CMAE restores the masked content of the same view, SIM reconstructs the features of another view. Our empirical results demonstrate that CMAE is not just simpler but also more effective in representation learning, as evidenced by superior performance.

\noindent\textbf{Input.} The majority of prior contrastive learning methods~\cite{he2020momentum,chen2020simple} implement robust augmentation techniques (e.g., random crop, random scale) to generate positive views from the same image. These operations are also commonly used in contrastive learning models under the masked image modeling scenario, such as the iBot. However, considering that the masking operation, which utilizes a large masking ratio (e.g. $75\%$ in~\cite{he2022masked}), already significantly degrades the input, applying these augmentations further could generate invalid positive views, thereby hindering contrastive learning. In contrast, we propose a novel, moderate data augmentation operation called pixel shifting for achieving better alignment between positive views. Compared to ExtreMA~\cite{wu2022extreme}, which employs the exact same view in two siamese branches, pixel shifting introduces a moderate input variance, which proves beneficial for contrastive learning (refer to Table~\ref{tab:abl_component}).

\noindent\textbf{Architecture.} In CMAE, a lightweight feature decoder is appended after the online encoder to supplement the masked features. This is a notable distinction from other methods, such as SIM and iBot, which directly utilize the representations of the visible patches to match that of the unmasked view. We contend that conducting contrastive learning between the features of the masked parts and the input images is impractical, considering they exhibit different levels of abstraction and semantic coverage. The feature decoder is anticipated to facilitate optimization by diminishing the distribution gap between contrastive features. The efficacy of the feature decoder is empirically validated, as shown in Table~\ref{tab:abl_decoder}. Notably, the design of CMAE is non-intrusive, allowing for its straightforward application to existing MIM models, such as MAE and ConvMAE, without necessitating significant modifications to the MIM model.

\begin{table*}[htp!]
    \centering
    \small
    \renewcommand\arraystretch{1.1}
    \begin{subtable}[h]{0.4\textwidth}
        \setlength{\tabcolsep}{10pt}
        \centering
        \begin{tabular}{lcccc} \toprule
        Method & Pre-Epochs & mIoU \\ \midrule
        MoCo-v3~\cite{chen2021empirical} & 300 & 47.3 \\
        DINO~\cite{caron2021emerging} & 400& 47.2 \\
        BEiT~\cite{bao2021beit} & 800  & 47.1\\
        CIM~\cite{fang2022corrupted} & 300 &  43.5 \\
        CAE~\cite{chen2022context} & 1600&  50.2 \\
        iBOT~\cite{zhou2021ibot} & 1600 & 50.0\\
        PeCo~\cite{dong2021peco} & 800 & 48.5\\
        MAE~\cite{he2022masked} & 1600 &48.1 \\
        CMAE & 1600  & \highlight{51.0} \\ \midrule
        ConvMAE\textsuperscript{*}~\cite{gao2022convmae} & 1600 & 50.7 \\
        CMAE\textsuperscript{*} & 1600 & \highlight{52.5} \\ \bottomrule
        \end{tabular}
       \caption{Semantic segmentation results on ADE20K. We use UperNet~\cite{xiao2018unified} as our default segmentation framework.}
       \label{tab:seg}
    \end{subtable}
    \hfill
    \begin{subtable}[h]{0.57\textwidth}
        \centering
        \setlength{\tabcolsep}{12pt}
        \begin{tabular}{lccccc}
    \toprule
         Method & Pre-Epochs  & AP$^{bbox}$ & AP$^{mask}$ \\
    \midrule
        MoCo-v3~\cite{chen2021empirical} & 300  & 47.9 & 42.7 \\
        BEiT~\cite{bao2021beit} & 800 &  49.8 & 44.4\\
        CAE~\cite{chen2022context} & 1600 & 50.0 & 44.0 \\
        iBOT$^\ddagger$~\cite{zhou2021ibot} & 1600 & 51.2 & 44.2\\
        PeCo~\cite{dong2021peco} & 800 & 44.9 & 40.4 \\
        SIM~\cite{tao2022siamese} & 1600 & 49.1 & 43.8 \\
        MAE$^\dagger$~\cite{he2022masked} & 1600 & 51.7 & 45.9 \\
        MAE~\cite{he2022masked} & 1600 & 50.3 & 44.9 \\
        CMAE & 1600 & \highlight{52.4} & \highlight{46.5}  \\
    \midrule
    
    ConvMAE\textsuperscript{*}~\cite{gao2022convmae} & 1600 & 52.5 & 46.5 \\
    CMAE\textsuperscript{*} & 1600 & \highlight{52.9} & \highlight{47.0} \\
    \bottomrule
        
    \end{tabular}
    \caption{COCO object detection and segmentation. We use the Mask R-CNN model~\cite{he2017mask} as our framework. $^\ddagger$ means using Cascade Mask R-CNN~\cite{cai2019cascade}.
        }
        \label{tab:coco_det}
     \end{subtable}
     \caption{Performance comparison on downstream tasks, including semantic segmentation
    and object detection. The symbol of \textsuperscript{*} denotes using convolutions to embed visual patches. $^\dagger$ denotes reproduced results of ours.}
     \label{tab:ablation}
\end{table*}

\section{Experiments}

\subsection{Implementation Details}

\noindent\textbf{Pre-training.} We follow the settings of MAE~\cite{he2022masked} to pre-train our model. We adopt AdamW~\cite{loshchilov2017decoupled} optimizer as default, and the momentum is set to $\beta_1 = 0.9$ , $\beta_2 = 0.95$. Besides, the weight decay is set to $0.05$. We use the linear scaling rule~\cite{goyal2017accurate}: $lr=base\_lr\times batch\_size/256$ to set the learning rate. The base learning rate is $1.5\times 10^{-4}$ with a batch size of $4096$. Cosine learning rate schedule~\cite{loshchilov2016sgdr} with a warmup of $40$ epochs is adopted. All pre-training experiments are conducted on $32$ NVIDIA A100 GPUs.

\noindent\textbf{Encoder Structure.} We use the ViT~\cite{dosovitskiy2020image} base model as our default setting. To further validate the extensibility of our proposed model, we replace the ViT with a hybrid convolutional ViT which is also used by ConvMAE~\cite{gao2022convmae}. In the hybrid ViT, a multi-layer convolutional network~\cite{lecun1989backpropagation} is used as token projection. Note the hybrid ViT is made to have the same model size as the ViT counterpart for fair comparison. We also experiment with scaled up encoders for evaluating the scalability of our method.
\begin{table*}
    \centering
    \begin{subtable}[h]{0.43\textwidth}
        \setlength{\tabcolsep}{15pt}
        \centering
        \begin{tabular}{l|c}
    Setting &  Accuracy\\
    \toprule
    Baseline~\cite{he2022masked}  & 82.9 \\
    + Contrastive learning & 83.1 \\
    + Pixel shifting aug. & 83.6 \\
    + Feature decoder & 83.8 \\
    
    \end{tabular}
       \caption{Component analysis.}
       \label{tab:abl_component}
      \vspace{10pt}
    \end{subtable}
    \hfill
    \begin{subtable}[h]{0.56\textwidth}
        \centering
        \setlength{\tabcolsep}{15pt}
        \begin{tabular}{ccc|c}
    
    Rand crop & Pixel shift & Color Aug. & Accuracy\\
    \toprule
    \xmark & \xmark & \xmark & 82.9 \\
    \cmark & \xmark & \xmark & 83.0 \\
    \xmark & \cmark & \xmark & 83.4 \\
    \xmark & \cmark & \cmark & 83.8 \\ 
    \end{tabular}
        \caption{Data augmentation analysis.}
        \label{tab:abl_augmentation}
        \vspace{10pt}
     \end{subtable}
    \bigskip 
    \begin{subtable}[h]{0.4\textwidth}
        \centering
        \setlength{\tabcolsep}{12pt}
        \begin{tabular}{c|cc}
            \#Blocks & Share weight & Accuracy\\
            \toprule
            0 & \xmark  & 83.6 \\
            2 & \xmark  & 83.8 \\
            2 & \cmark  & 83.4 \\
            4 & \xmark  & 83.8 \\
            4 & \cmark  & 83.5 \\
        \end{tabular}
        \caption{Feature decoder analysis.}
        \label{tab:abl_decoder}
    \end{subtable}
    \hfill
    \begin{subtable}[h]{0.29\textwidth}
        \centering
        \setlength{\tabcolsep}{12pt}
        \begin{tabular}{c|c}
            Loss weight & Accuracy\\
            \toprule
            0.1 &  83.3 \\
            0.5 &  83.7 \\
            1.0 &  83.8 \\   
            1.5 &  83.5 \\
            2.0 &  83.2 \\
        \end{tabular}
        \caption{Contrastive loss weight.}
        \label{tab:abl_contra_loss_weight}
     \end{subtable}
     \hfill
     \begin{subtable}[h]{0.3\textwidth}
        \centering
        \setlength{\tabcolsep}{12pt}
        \begin{tabular}{c|c}
            Masking ratio & Accuracy \\
            \toprule
            0 &  83.8 \\
            0.25 &  83.6 \\
            0.5 &  83.3 \\   
            0.65 &  83.3 \\
            0.75 &  83.0 \\
        \end{tabular}
        \caption{ Momentum encoder masking ratio.}
        \label{tab:abl_mask_ratio}
     \end{subtable}
     \vspace{-10pt}
    \caption{Ablations. We evaluate all models on ImageNet-1k with their top-1 classification accuracies. Each model is pre-trained for $300$ epochs.}
    \label{tab:abl_sum}
\end{table*}

\subsection{Results on ImageNet}
Following existing works~\cite{chen2021empirical,bao2021beit,xie2022simmim,he2022masked}, we use ImageNet-1K~\cite{deng2009imagenet} which consists of 1.3M images of 1k categories as the pre-training and fine-tuning dataset. The dataset contains two subsets: the training set and the validation set. We only use the training set to pre-train CMAE. After pre-training, the CMAE online encoder is used for fine-tuning on ImageNet-1k training set for $100$ epochs. For the model pre-trained with $300$ epochs, we adopt $5.e^{-4}$ as the base learning rate in fine-tuning. Since the longer pre-training schedule ($1600$ epochs) makes the model learn better initialization weights for fine-tuning~\cite{xie2022data}, we set a smaller base learning rate of $2.5e^{-4}$. Besides, we follow the common fine-tuning practices to regularize the model using mixup~\cite{zhang2018mixup}, cutmix~\cite{yun2019cutmix}, drop path~\cite{huang2016deep}, etc. 

In Table~\ref{tab:ImageNet-1k}, we compare CMAE with competing methods on the fine-tuning classification accuracy on ImageNet. CMAE achieves a top-1 accuracy of $84.7\%$, which is $1.1\%$ higher than MAE~\cite{he2022masked}. Among all models using ViT architecture, CMAE achieves the best performance. Compared with contrastive learning based methods Moco-v3~\cite{chen2021empirical} and DINO~\cite{caron2021emerging}, our model can significantly outperform them by $1.5\%$ and $1.9\%$ respectively. Compared with iBOT and SIM which also use contrastive objective in MIM, our CMAE achieves higher performance with a gain of $0.7\%$ and $0.8\%$, respectively. Above results strongly evidence the superiority of CMAE.

When we replace the vanilla ViT encoder with a hybrid convolutional ViT, as done in ConvMAE~\cite{gao2022convmae}, CMAE further improves to $85.0\%$ and $85.3\%$ with the pre-training of $800$ epochs and $1600$ epochs, respectively. These results surpass those of ConvMAE under the same pre-training setting by $0.4\%$ and $0.7\%$ respectively, verifying the excellent extendibility of CMAE to various network structures.

Remarkably, CMAE can gain a noticeable improvement with a prolonged training schedule (from $800$ epochs to $1600$ epochs) while ConvMAE is observed to saturate at $800$ epochs. This result suggests the stronger capability of CMAE on learning better representations.

\subsection{Transfer Learning}

To further validate the transferability of CMAE, we follow previous methods to evaluate pre-trained models on the semantic segmentation dataset ADE20K~\cite{ade20k}, the object detection dataset COCO2017~\cite{lin2014microsoft} and classification datasets.

\noindent\textbf{Semantic segmentation.} ADE20K~\cite{ade20k} has 25,562 images of $150$ fine-grained categories. We adopt Upernet~\cite{xiao2018unified} as the default model for this task, following the settings of compared methods. 
The backbone ViT-B is initialized from pre-training while other modules are initialized with the Xavier~\cite{glorot2010understanding} initialization. The model is fine-tuned on the training set of ADE20K and tested on standard validation split.

Following previous works, we report the Mean Intersection over Union~(mIoU) performance of CMAE in Table~\ref{tab:seg}. We notice that CMAE significantly surpasses MAE by $2.9\%$, which verifies the stronger transferability of CMAE. Besides, CMAE also improve by $1.0\%$ and $0.8\%$ compared with iBOT~\cite{zhou2021ibot} and CAE~\cite{chen2022context} respectively. With the same hybrid ViT backbone, CMAE significantly outperforms ConvMAE by $1.8\%$. Remarkably, CMAE sets a new state-of-the-art result of $52.5$ by surpassing all competing methods with a large margin.

\noindent\textbf{Object Detection and Segmentation.} We adopt the widely used object detection and instance segmentation framework Mask-RCNN~\cite{he2017mask,li2021benchmarking} for benchmarking on this task. ViT-B is used as the backbone and initialized with our pre-trained model. Following MAE, we fine-tune the model on COCO train2017 split, and report box AP for object detection and mask AP for instance segmentation on val2017 split. We fine-tune the model for $100$ epochs. The base learning rate is $1.e^{-4}$ with a cosine annealing schedule, and the weight decay is set to $0.1$.

The comparison results with other self-supervised learning methods are shown in Table~\ref{tab:coco_det}. As one can see, CMAE improves over MAE from $51.7$ to $52.4$ on AP\textsuperscript{b} and from $45.9$ to $46.5$ on AP\textsuperscript{m}. With the hybrid ViT structure, CMAE consistently surpasses the competing method ConvMAE: AP\textsuperscript{b} increases from $52.5$ to $52.9$ and AP\textsuperscript{m} increases from $46.5$ to $47.0$. Above promising results again verify the effectiveness of our method.

\begin{table}[]
    \centering
    \begin{tabular}{c|c|c|c|c}
       \toprule
        Method &  iNat2017 & iNat2018 & iNat2019 & Places365 \\
        \midrule
        MAE    &   70.5    &    75.4  & 80.5     & 57.9     \\
        CMAE   &   72.2    &    76.4  & 82.2     & 58.9      \\
        \bottomrule
    \end{tabular}
    \caption{Transfer learning accuracy on classification datasets.}
    
    \label{tab:classification_acc}
\end{table}
\noindent\textbf{Classification tasks.} To further study transfer learning on classification tasks, we validate our model on the iNaturalists~\cite{van2018inaturalist} and Places~\cite{zhou2014learning} in Table~\ref{tab:classification_acc}. Experiments across four classification tasks on these datasets showed consistent improvements of $1.0\%$ to $1.7\%$ in top-1 accuracy over the MAE~\cite{he2022masked}. These results provide further evidence for the efficacy of our method in enhancing the discriminative capabilities of pretaining model.

\subsection{Method Analysis}
\label{ablation}

To understand the effects of key components and validate design choices we adopt in CMAE, we conduct a series of ablation experiments. Unless otherwise stated, we report the performance of our model with $300$ pre-training epochs in this subsection. The ablative results are listed in Table~\ref{tab:abl_sum}. In the following, we verify the effectiveness of our main design ideas, then conduct ablation experiments for each component separately.

\noindent\textbf{Ablation of components.} In Table~\ref{tab:abl_component}, we show how each component, i.e. contrastive learning, pixel shifting data augmentation and feature decoder affects model's performance. We start with a vanilla implementation of contrastive learning on MAE. Specifically, following the input generation approach in contrastive methods, random cropped regions with masking are fed into online/momentum encoder. The same contrastive objective as introduced in Section~\ref{loss_function} are optimized between the output of online encoder and momentum encoder. As can be seen from Table~\ref{tab:abl_component}, such an intuitive approach only leads to marginal improvement ($0.2\%$). Apparently, the power of contrastive learning is not fully unleashed due to ignoring its compatibility with MIM. By using the proposed moderate data augmentation, i.e. pixel shifting, the result can increase from $83.1\%$ to $83.6\%$, which evidences the advantage of pixel shifting. Moreover, applying feature decoder further boosts the model's learning capability by improving the performance to $83.8\%$, demonstrating its effectiveness in our method.

\noindent\textbf{Contrastive loss.} To explore the effect of contrastive loss in CMAE, we experiment with various loss weights, i.e. $\lambda_c$ in Eq.~\eqref{eq:total_loss}. The results are shown in Table~\ref{tab:abl_contra_loss_weight}. Note CMAE degenerates to the baseline MAE when loss weight is 0. When increasing the weights from $0$ to $1$, the model's performance increases accordingly, which verifies the importance of contrastive learning on enhancing the learned representations. When the weight of contrast learning is greater than that of MIM, we observe the phenomenon of imbalanced training occurs which adversely affects the final performance. This experiment demonstrates that both contrastive loss and reconstructive loss are critical for learning capable representations. We therefore set $\lambda_c=1$ throughout our experiments.

We also conduct controlled experiments with different contrastive loss forms to compare their influences on pre-training. Under the same configuration, we observe that the model trained with InfoNCE loss achieves higher performance than BYOL-style loss ($83.8\%$ vs. $83.4\%$). This result suggests that the way of utilizing negative samples in InfoNCE is more effective in our method.

\noindent\textbf{Pixel shifting augmentation.} In this section, we ablate on the importance of data augmentations. In contrast to the common practices of applying heavy data augmentation in contrastive learning, we find a moderate data augmentation is more effective in aligning contrastive learning and MIM. 
We divide data augmentation methods into two kinds: spatial transfer and color transfer, and evaluate their effect respectively. For spatial transfer, we compare our proposed pixel shifting with the commonly used randomly resized cropping. For color transfer, we compare two cases, i.e. with or without using color jittering for the momentum branch. 

\begin{table}[]
    \centering
    \resizebox{\columnwidth}{!}{ 
    \begin{tabular}{c|cccccccc}
    \toprule
    Shift value &random crop & 0 & 0-15 & 0-31 & 0-47 & 0-63 & 32-47 & 48-63 \\
    \midrule
    Accuracy &83.29& 83.68 & 83.78 & 83.82 & 83.71 & 83.64 & 83.54  & 83.48 \\
    \bottomrule
    \end{tabular}
    } 
    \caption{ Impact of varying pixel shift ranges on the ImageNet-1k classification task. ``Random crop'' serves as the baseline method for comparison against our proposed pixel shift approach.}
   
    \label{tab:ablation_shift_value}
\end{table}

\begin{figure*}
    \centering
    \begin{minipage}{0.48\linewidth}
        \includegraphics[width=\linewidth]{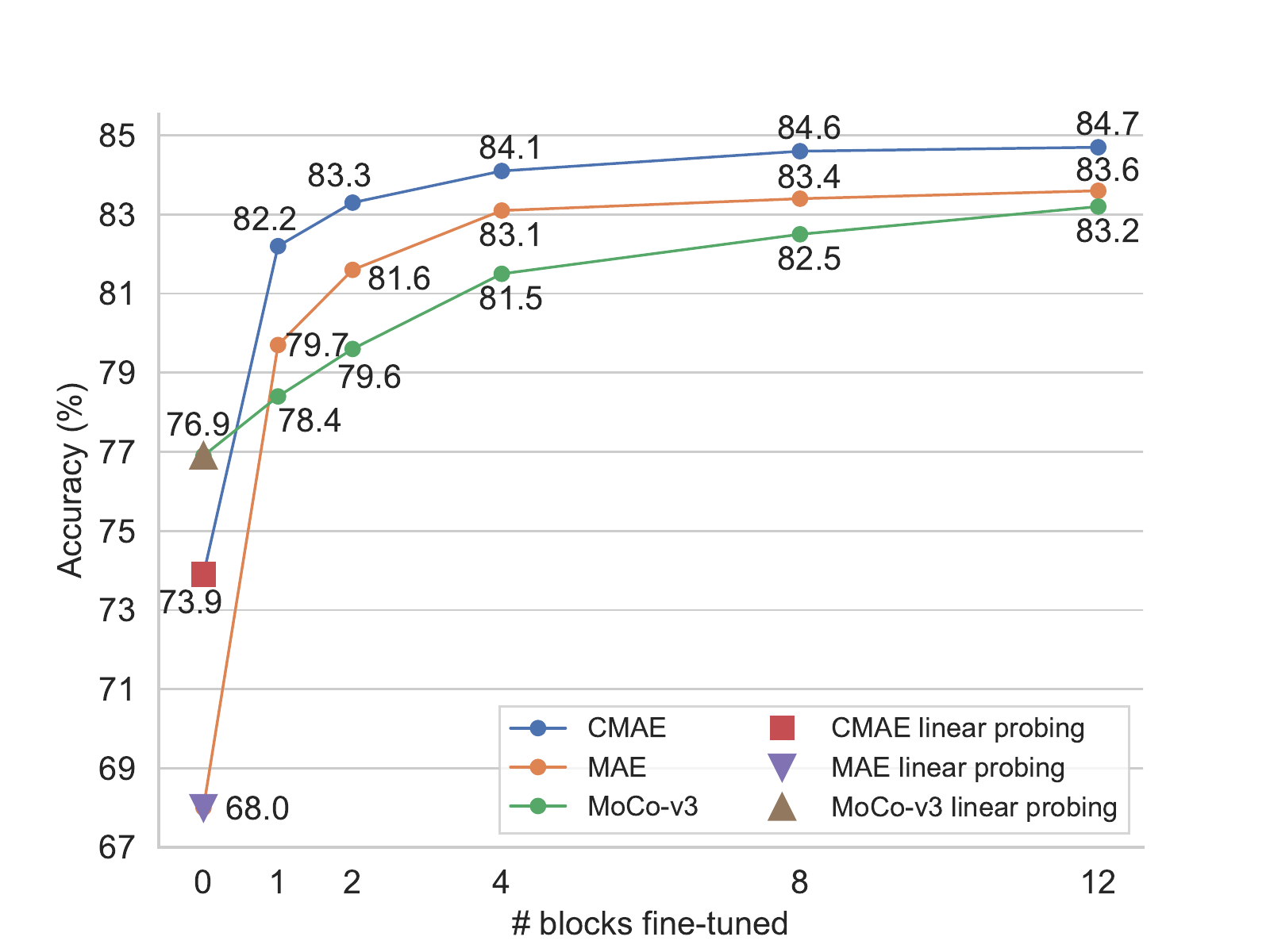}
        \subcaption{Partial fine-tuning results using the ViT-B backbone.}
        \label{fig:partial_fine-tuning}
    \end{minipage}
    \hfill
    \begin{minipage}{0.48\linewidth}
        \includegraphics[width=\linewidth]{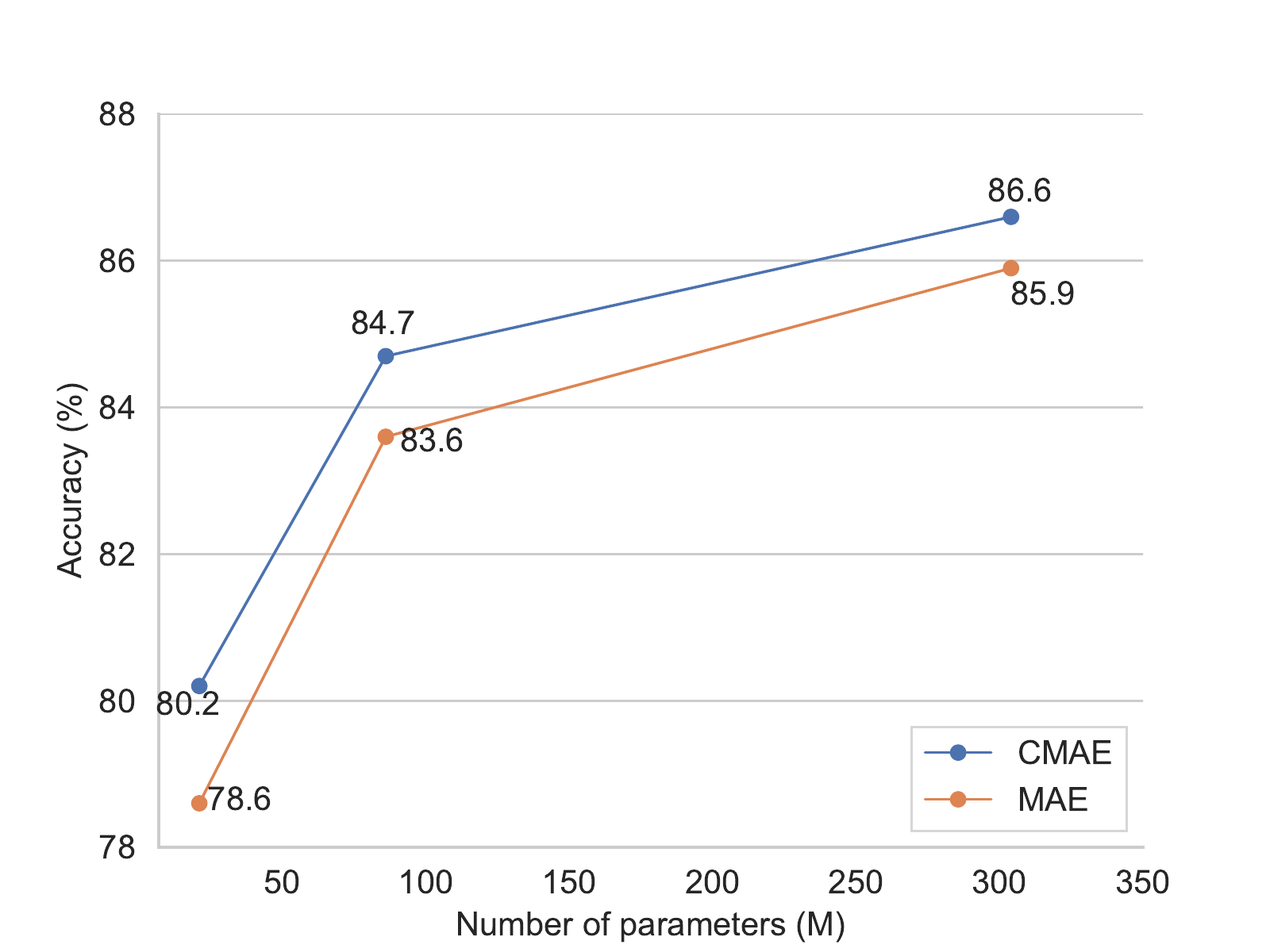}
        \subcaption{Scaling
        results with different model sizes. }
        \label{fig:model_scale_up}
    \end{minipage}
    \caption{Performance comparison on ImageNet-1k with partial fine-tuning and model scaling. In partial fine-tuning, we adopt the model weights pre-trained with $1600$ epochs. In model scaling experiments, all models are pre-trained with $1600$ epochs.}
\end{figure*}

As can be seen from Table~\ref{tab:abl_augmentation}, pixel shifting significantly surpasses random crop ($83.4\%$ vs. $83.0\%$). The superiority of pixel shifting should be attributed to its ability of generating more plausible positive views. As introduced in Section~\ref{sec:view_aug}, this property helps contrastive learning better collaborate with MIM in our framework. By using color transfer, the result further improves to $83.8\%$, suggesting color transfer is complementary to our method.

We investigate the impact of different pixel shift ranges by varying the maximum allowable shift. Intuitively, larger shift ranges introduce greater misalignment between the two augmented views. As evident in Table~\ref{tab:ablation_shift_value}, excessive shifts severely degrade model performance, conforming to our hypothesis that severely misaligned positive pairs introduce noise detrimental to contrastive learning. The results demonstrate an optimal balance exists between view diversity and alignment. Based on the observed performance across different shift ranges, we select a maximum pixel shift of 31 as the default parameter setting to maximize contrastive learning while preserving sufficient alignment.

\noindent\textbf{Feature decoder.} Different from existing works,  we introduce a feature decoder to recover the features of masked patches when performing contrastive learning. To investigate its effectiveness, we present experiments under following two settings: sharing the weight between feature decoder and pixel decoder or not, and changing the depth of feature decoder.

\begin{figure}
    \centering
    \includegraphics[width=0.88\linewidth]{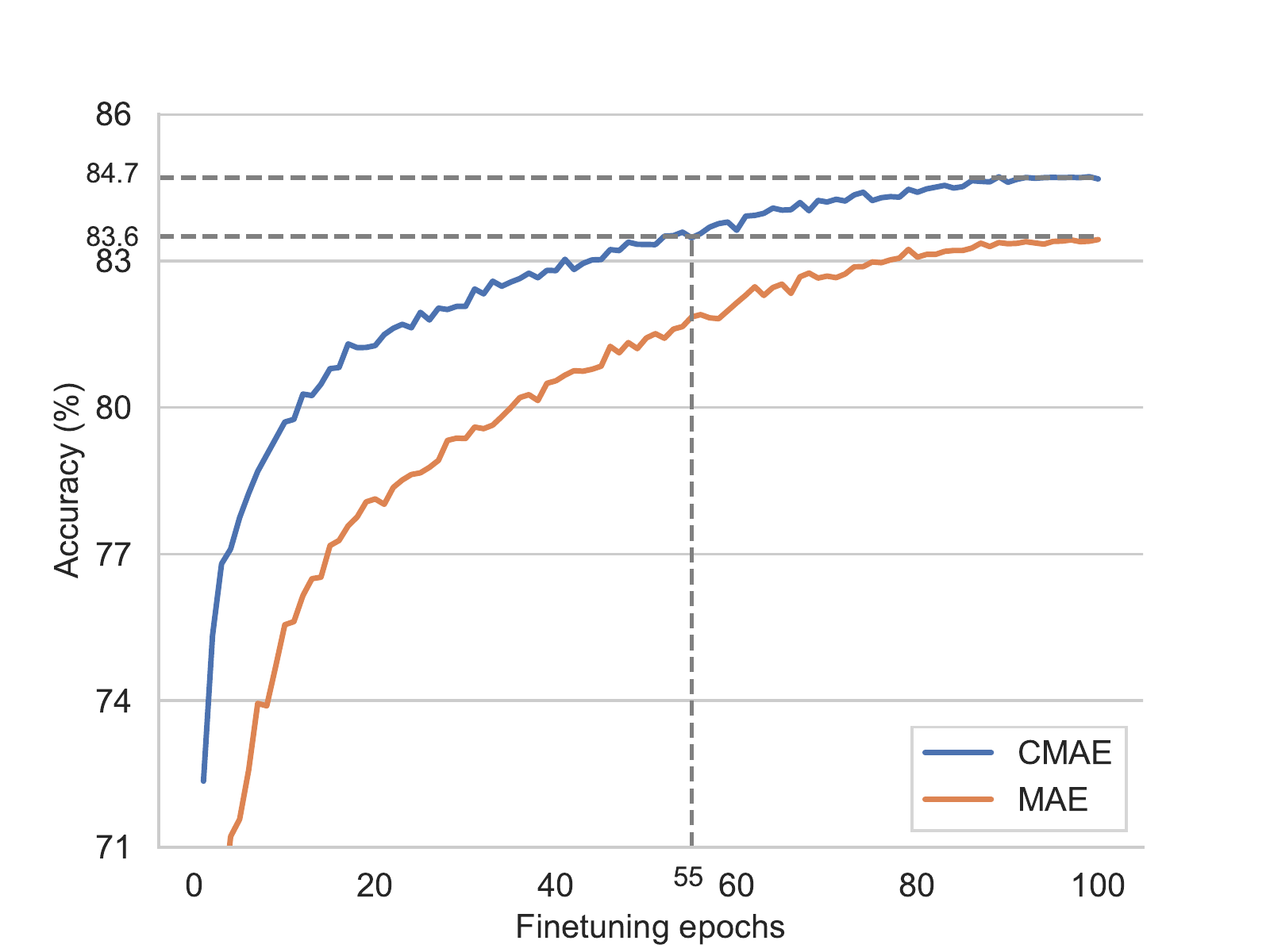}
    \caption{Convergence speed compared with MAE. All models are pre-trained with $1600$ epochs.  }
        \label{fig:ablation_convergence}
\end{figure}

In Table~\ref{tab:abl_decoder},   number ``0'' means not using feature decoder, i.e. the output of the online encoder which contains only the features of visible tokens is used for contrastive learning. Under this setting, our method performs worse than using a lightweight two-layer feature decoder. When increasing the depth of feature decoder, there is no significant impact on performance. However, when the depth increases to $8$, we obtain a trivial solution, possibly due to the optimization difficulty caused by deeper structure. To strike a balance between efficiency and effectiveness, we set the depth to be $2$.
Besides, when the feature decoder shares the weight with the pixel decoder, the method performs the worst. A plausible explanation is that the two branches have different targets, thus should adopt independent weights.

\noindent\textbf{Masking ratio for the momentum branch.} In this experiment, we investigate whether masking a portion of image patches for the momentum branch affects the model performance.
Following previous works, we select a set of masking ratios, including $\{0, 0.25, 0.5, 0.65, 0.75\}$ for the momentum branch and see how the performance changes.
As shown in Table~\ref{tab:abl_mask_ratio}, one can observe that using the complete set of the image tokens yields the best results.
A possible reason is that: since the aim of adding the momentum branch is to provide our model with the contrastive supervision, incorporating the full semantics of an image is preferred.
Otherwise, the masked input with degenerated semantic information may lead to a sub-optimal solution in contrastive learning. Based on this observation, the momentum branch in our model uses the whole image as input throughout our experiments. 

\begin{figure*}
    \centering
    \begin{minipage}{0.33\linewidth}
        \includegraphics[width=\linewidth]{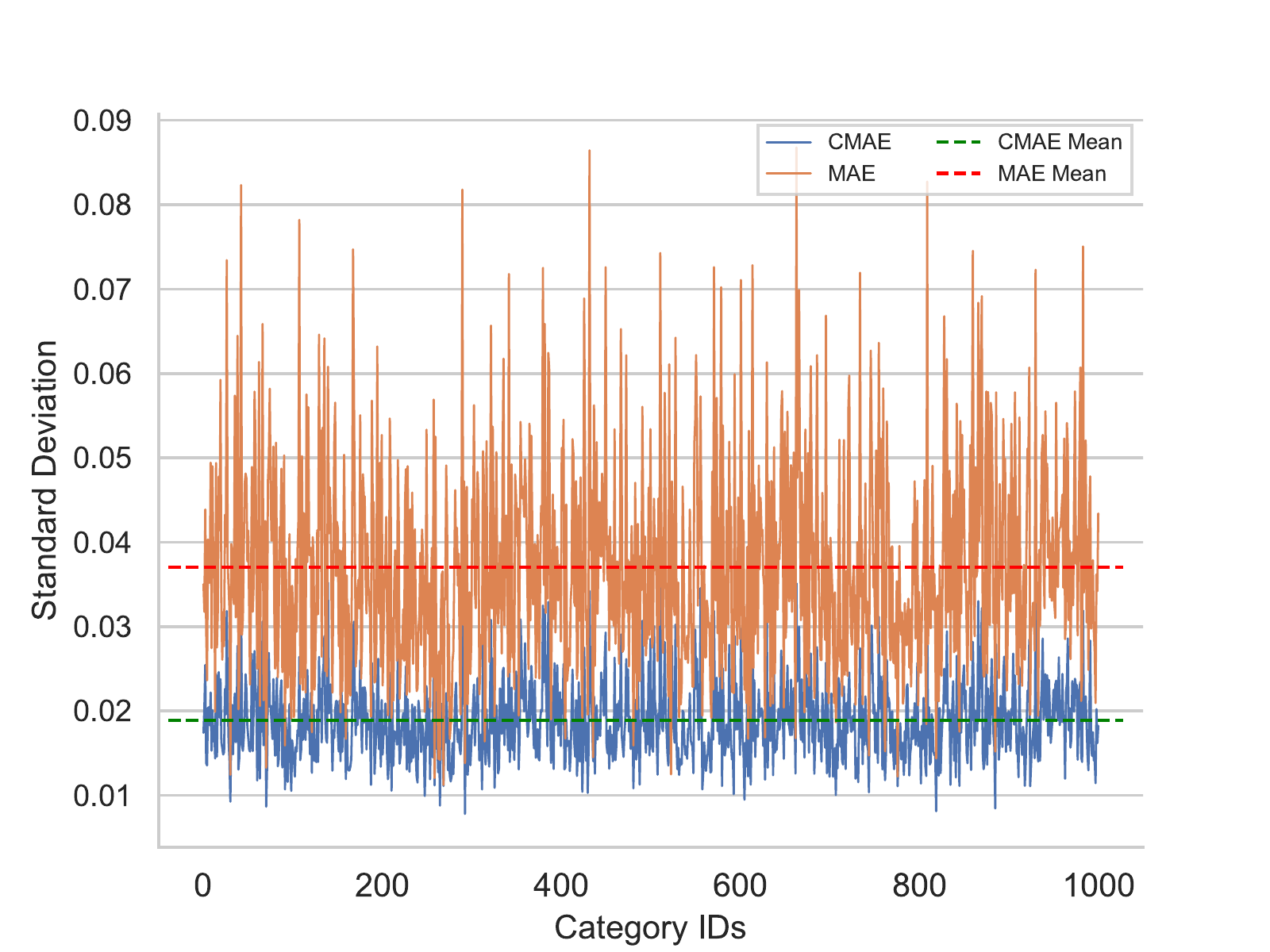}
        \subcaption{Standard variance of intra-class distance. }
        \label{fig:intra-class_std}
    \end{minipage}
    \begin{minipage}{0.33\linewidth}
        \includegraphics[width=\linewidth]{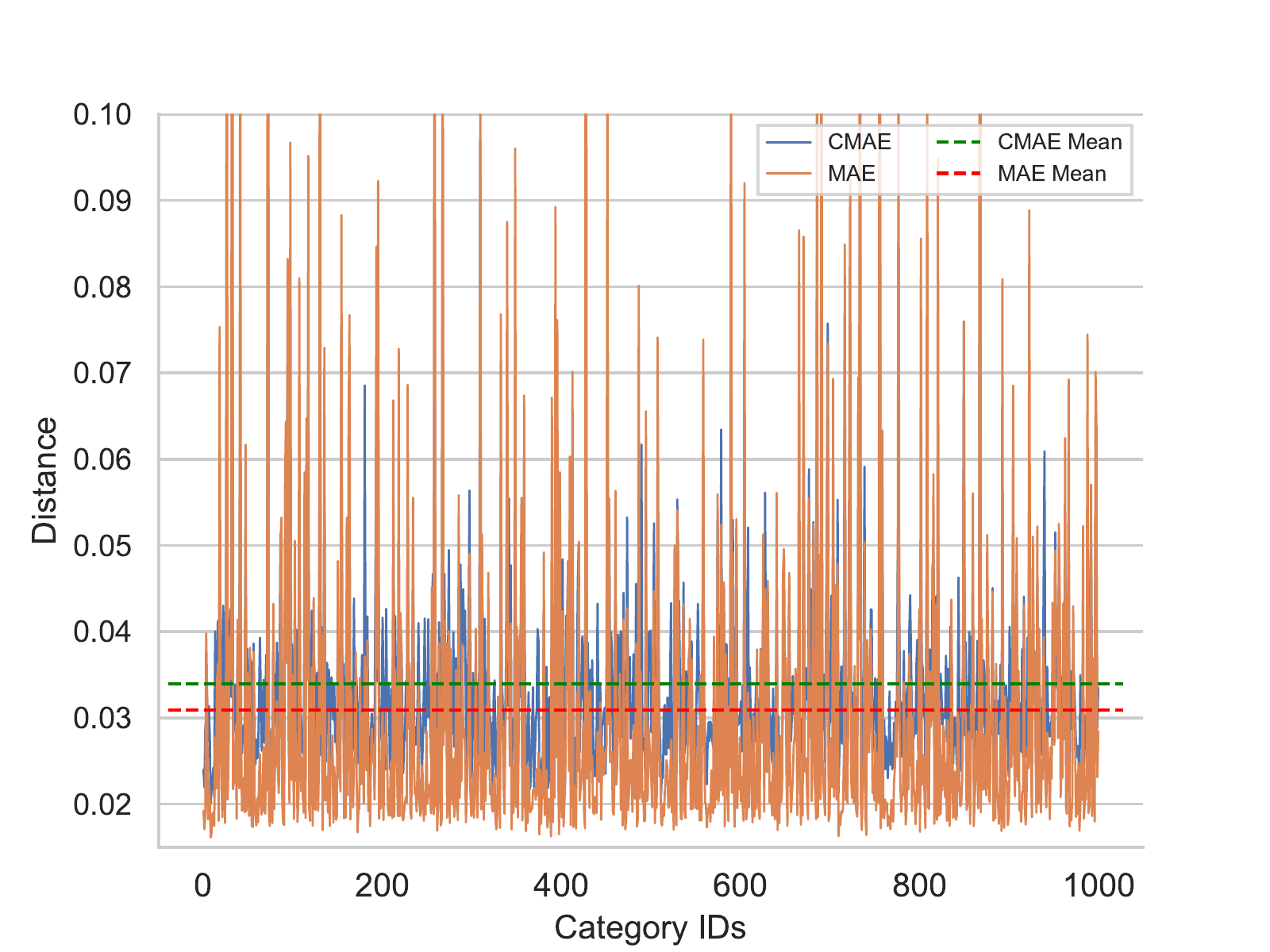}
        \subcaption{Average distance of inter-class center}
        \label{fig:inter-class_ava}
    \end{minipage}
    \begin{minipage}{0.33\linewidth}
        \includegraphics[width=\linewidth]{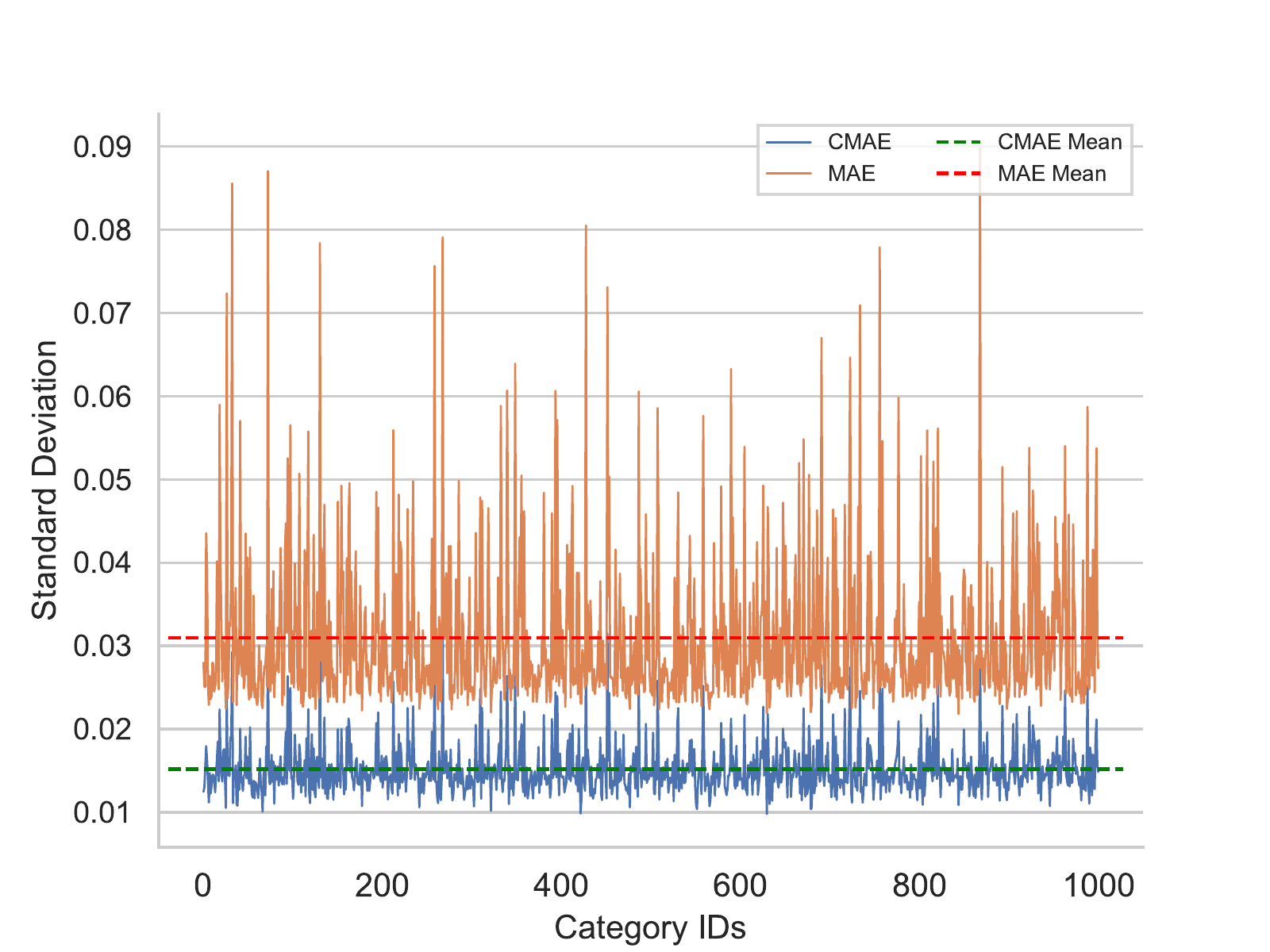}
        \subcaption{ Standard variance of inter-class center distance. }
        \label{fig:inter-class_std}
    \end{minipage}
    
    \caption{Analysis of features of pre-training models. In terms of intra-class distance, our model achieves a lower standard deviation relative to MAE on the ImageNet-1k validation set. With respect to inter-class center distance, our model displays greater inter-class distances than MAE for 811 out of the 1000 categories, and manifests a reduced standard deviation of distances among category centers. }
    \label{fig:feature_std_va}
\end{figure*}

\noindent\textbf{Convergence speed.} To further show our method's effectiveness, we compare the convergence behavior of CMAE and MAE when fine-tuning on ImageNet-1k. The pre-trained weights with $1600$ epochs are used as initialization. As shown in Figure~\ref{fig:ablation_convergence}, we observe that CMAE converges much faster compared with MAE: with only $55$ fine-tuning epochs, CMAE already surpasses the final performance of MAE. This result demonstrates that the representations learned by CMAE can be more easily adapted for specific tasks, an appealing property which is in line with the purposes of self-supervised pre-training.

\vspace{-5pt}
\subsection{Partial Fine-tuning and Linear Probing}
\vspace{-5pt}

In the context of task-specific training, both partial fine-tuning~\cite{zhang2016colorful,he2022masked,yosinski2014transferable,noroozi2016unsupervised} and linear probing methodologies retain the majority of the model components in a frozen state. However, a key distinction lies in the nature of the head being tuned: partial fine-tuning employs a non-linear head, whereas linear probing utilizes a linear one. As underscored by~\cite{he2022masked}, given that linear probing exhibits minimal correlation with transfer learning performance, partial fine-tuning emerges as a superior protocol for the evaluation of non-linear, yet more potent, representations. In light of these observations, our study also places emphasis on the partial fine-tuning metric.

Specifically, we follow the experimental settings of~\cite{he2022masked} to ablate the CMAE base model with $1600$ epoch pre-training. As shown in Figure~\ref{fig:partial_fine-tuning}, the performances of our model are consistently better than MAE in all tested settings, e.g. when fine-tuning one block, we get a $2.5\%$ gain over MAE. Above results demonstrate that our model can effectively improve the representation quality of baseline method. Note when the number of fine-tuned blocks is ``0'', it degenerates to linear probing. In this case, our model achieves significant improvement ($5.9\%$) over MAE. These results indicate that our method is able to improve the representation quality under both evaluation metrics. Furthermore, in comparison to the typical contrastive model MoCo-V3~\cite{chen2021empirical}, MoCo-V3 exhibits superior performance in the linear probing setting. However, under the partial fine-tuning setting, CMAE surpasses MoCo-V3 in all aspects, particularly when fine-tuning only one block, where CMAE yields a $3.8\%$ enhancement. This also serves as evidence that the features learned by our model are of higher quality.

\vspace{-5pt}
\subsection{Model Scaling}
\vspace{-5pt}

To study the scalability of our method for models of different sizes, we adopt ViT-small, ViT-base, and ViT-large as encoders and report their performance on ImageNet-1k fine-tuning. As shown in Figure~\ref{fig:model_scale_up}, CMAE can consistently boost the performance of MAE at all scales. These results clearly demonstrate the excellent scalability of CMAE.

\subsection{Feature Analysis}
In order to more effectively scrutinize the features obtained by our model, we utilize the ViT-base model for our investigation. Following the completion of pre-training, we extract features from the ImageNet-1k validation set and compute the following metrics:

\begin{itemize}

\item Average intra-class distance: This metric measures the mean distance between all pairs of images within the same class.

\item Standard deviation of intra-class distances: This metric measures the variation in distances between images in the same class.

\item Average inter-class distance: This metric measures the mean distance between all pairs of images from different class centers.

\item Standard deviation of inter-class distances: This metric measures the variation in distances between images from different class centers.

\end{itemize}

When we compute the average intra-class distance, CMAE achieves a lower average intra-class distance of 0.0377 than MAE's 0.0380. Furthermore, as shown in Figure~\ref{fig:intra-class_std}, CMAE has a smaller standard deviation of intra-class distances than MAE~(0.0189 vs. 0.0371). These results suggest that the features extracted by CMAE are more compactly clustered in the latent space. Regarding inter-class distances, we compute the average distance of each class center to other class centers and the standard deviation of distances to other class centers. As shown in Figures~\ref{fig:inter-class_ava} and~\ref{fig:inter-class_std}, CMAE demonstrates larger average inter-class distances (0.0340 vs. 0.0309) and a smaller standard deviation for inter-class distances (0.0152 vs. 0.0310). This indicates that the features extracted by CMAE have a more uniform distribution for each category in the latent space and larger inter-class distances.

In summary, the aforementioned results provide evidence that our model is capable of learning superior visual representations with enhanced discriminability compared to MAE.

\vspace{-5pt}
\section{Conclusion}
\vspace{-5pt}
This paper introduces a novel self-supervised learning framework named contrastive masked autoencoder (CMAE) which aims to improve the representation quality of MIM by leveraging contrastive learning. In CMAE, we propose two novel designs from the perspective of input generation and architectures respectively to harmonize MIM and contrastive learning. Through extensive experiments, it is demonstrated that CMAE can significantly improve the quality of learned representation in pre-training. Notably, on three well-established downstream tasks, i.e. image classification/segmentation/detection, CMAE achieves state-of-the-art performance. In the future, we will investigate the scaling up of CMAE to larger datasets and incorporate image-dense caption as another view for contrastive learning training based on CMAE.

\bibliographystyle{abbrvnat}
\bibliography{mybib_jr}

\ifCLASSOPTIONcaptionsoff
  \newpage
\fi

 \vspace{-30pt} 
\begin{IEEEbiography}[{\includegraphics[width=1in,height=1.25in,clip,keepaspectratio]{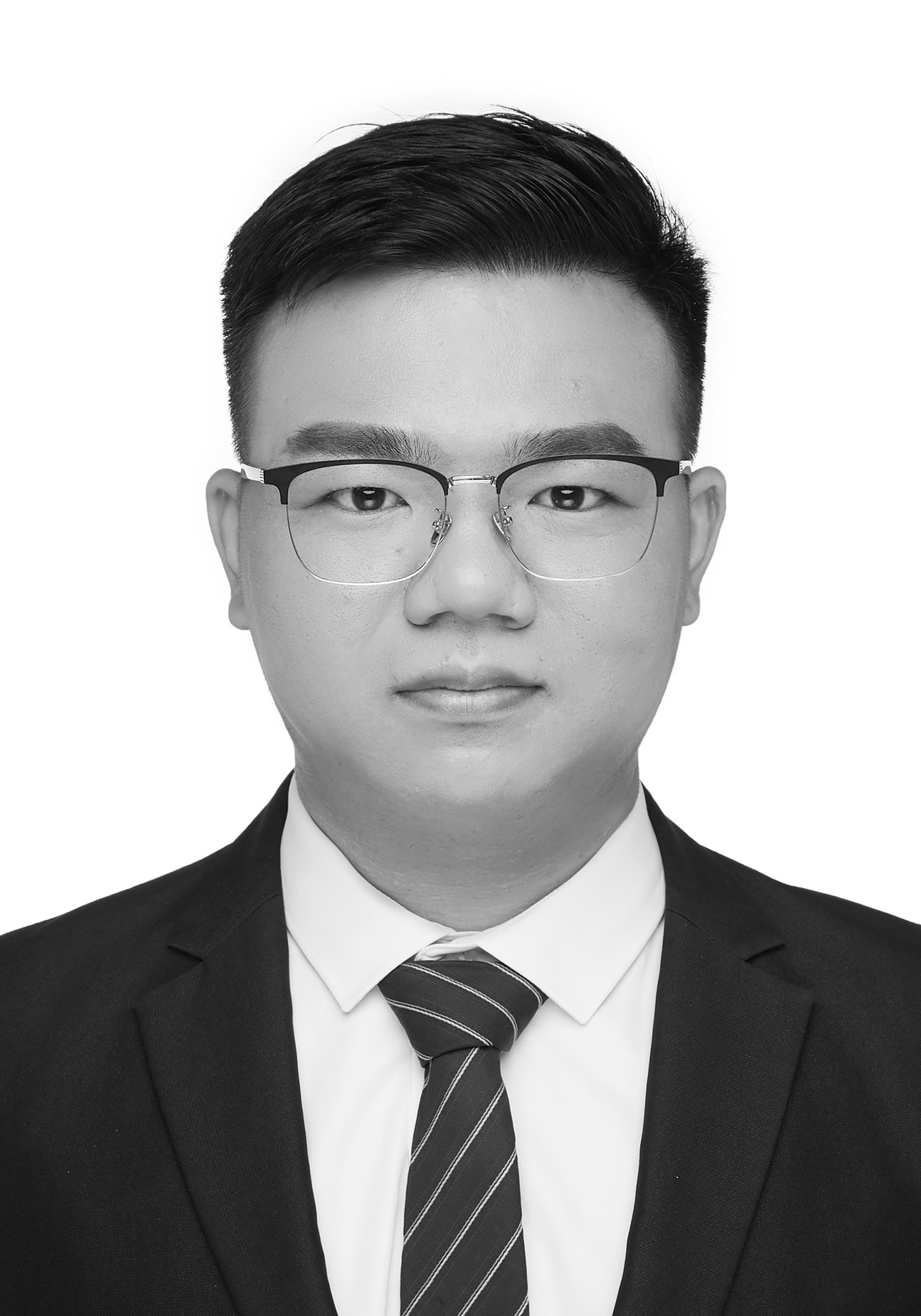}}]
{Zhicheng Huang} received the Master degree in artificial intelligence from University of Science and Technology Beijing, China, in 2019. He is pursuing Ph. D. at the School of  Automation and  Electrical  Engineering.  His research interests include self-supervised learning, visual-language pre-training and computer vision.
\end{IEEEbiography}
 \vspace{-30pt} 
\begin{IEEEbiography}[{\includegraphics[width=1in,height=1.25in,clip,keepaspectratio]{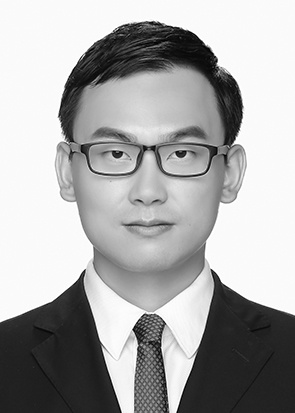}}]
{Xiaojie Jin} is working as a Research Scientist in Bytedance Inc. in the USA. He received his Ph.D degree from National University of Singapore in 2018. His current research interests include self-supervised learning, multi-modal understanding, intelligent video editing and efficient deep model. His research works have been published in top-tier conferences/journals, including ICCV, ECCV, CVPR, ICML, NeurIPS, ICLR, TPAMI, TNNLS, etc.
\end{IEEEbiography}
 \vspace{-30pt} 
\begin{IEEEbiography}[{\includegraphics[width=1in,height=1.25in,clip,keepaspectratio]{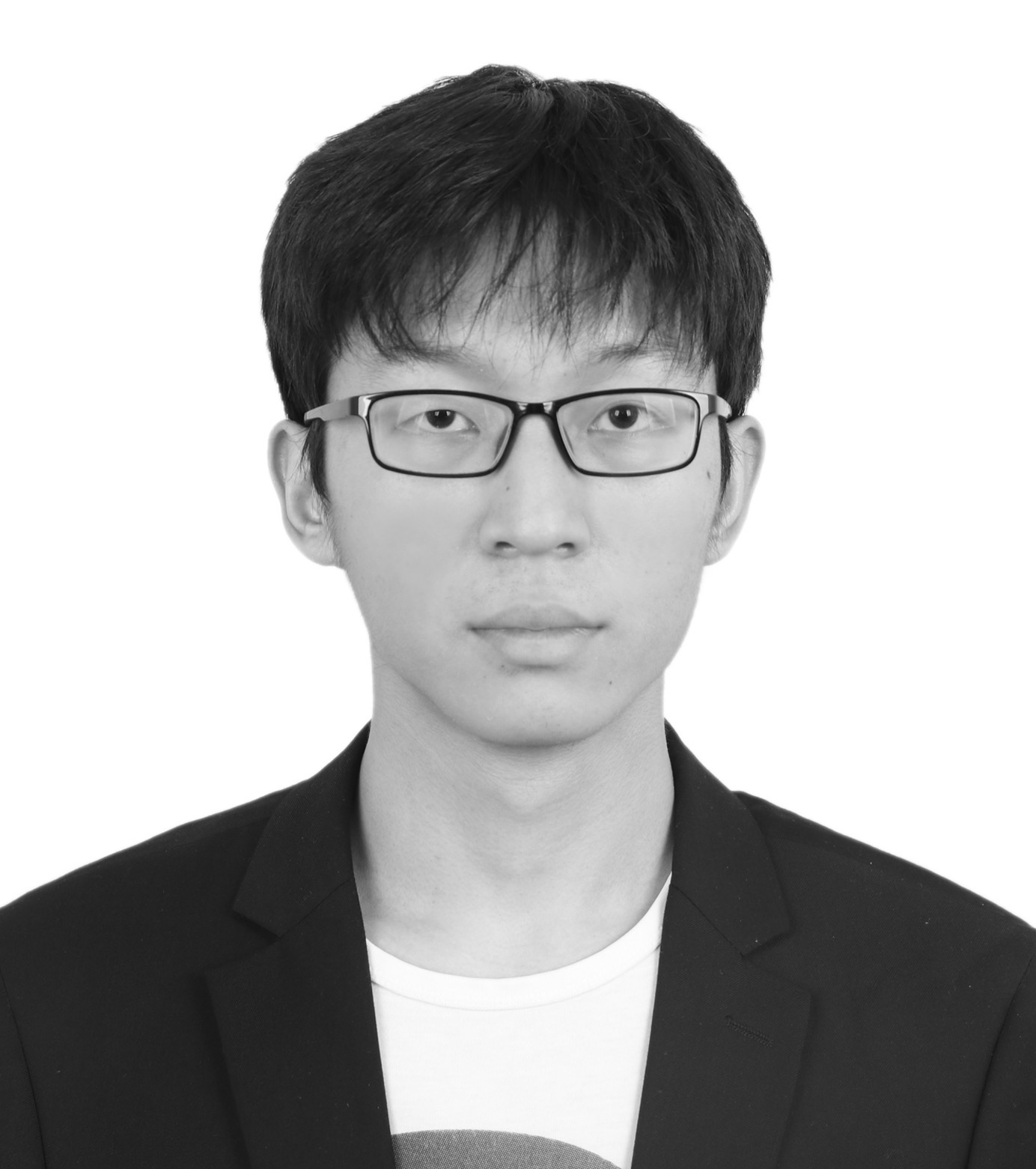}}]
{Cheng-Ze Lu} is currently a master student from the College of Computer Science at
Nankai University, under the supervision of Prof. Ming-Ming Cheng. 	
Before that, he received his B.E. degree from Xidian University in 2020.
His research interests include deep learning and computer vision.

\end{IEEEbiography}
 \vspace{-30pt} 
\begin{IEEEbiography}[{\includegraphics[width=1in,height=1.25in,clip,keepaspectratio]{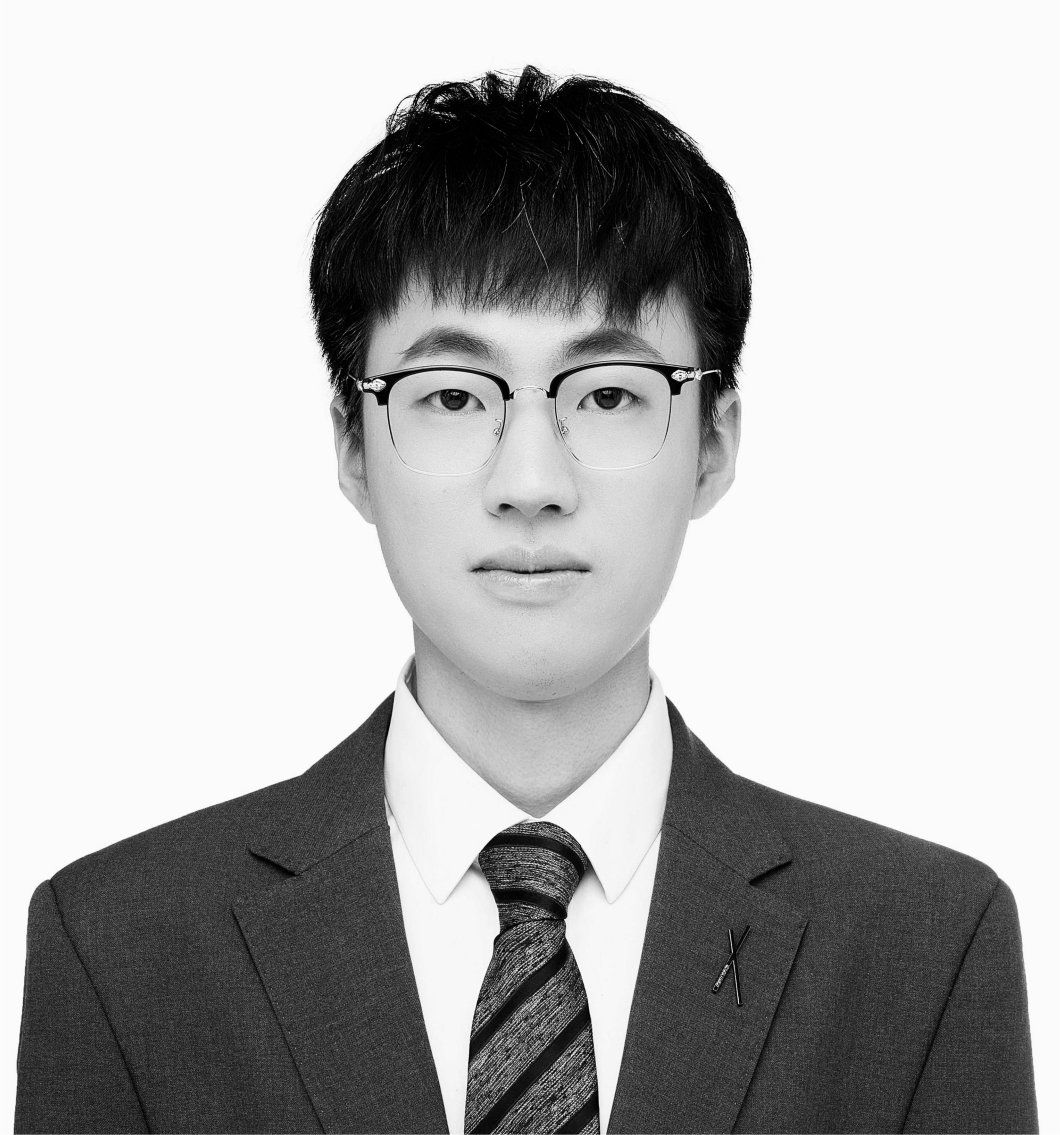}}]
{Qibin Hou} received his Ph.D. degree from the School of Computer Science, Nankai University. Then, he worked at the National University of Singapore as a research fellow. Now, he is an associate professor at School of Computer Science, Nankai University. He has published more than 20 papers on top conferences/journals, including T-PAMI, CVPR, ICCV, NeurIPS, etc. His research interests include deep learning, image processing, and computer vision.
\end{IEEEbiography}

 \vspace{-30pt} 
 
\begin{IEEEbiography}[{\includegraphics[width=1in,height=1.25in,clip,keepaspectratio]{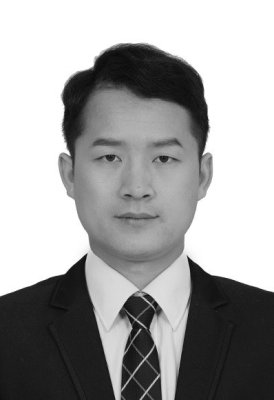}}]
{Ming-Ming Cheng} received his PhD degree from Tsinghua University in 2012.
Then he did 2 years research fellow, with Prof. Philip Torr in Oxford.
He is now a professor at Nankai University, leading the Media Computing Lab.
His research interests include computer graphics, computer vision, 
and image processing. 
He received research awards including ACM China Rising Star Award, 
IBM Global SUR Award, CCF-Intel Young Faculty Researcher Program.
\end{IEEEbiography}
 \vspace{-30pt} 
\begin{IEEEbiography}[{\includegraphics[width=1in,height=1.25in,clip,keepaspectratio]{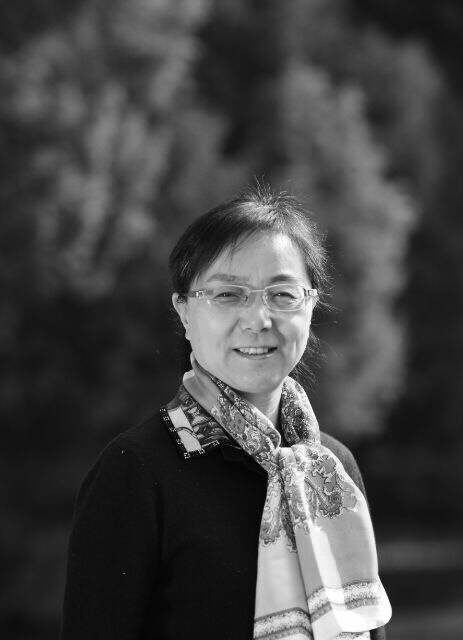}}]
{Dongmei Fu} received the M.S. degree from Northwestern Polytechnical University, in 1984, and the Ph.D. degree in automation science from the University of Science and Technology Beijing (USTB), China, in 2006, where she is currently a Professor and a Doctoral Supervisor. She has taken charge of several national projects about corrosion data mining and infrared image processing. Her current research interests include automation control theory, image processing, and data mining.
\end{IEEEbiography}
 \vspace{-30pt} 

\begin{IEEEbiography}[{\includegraphics[width=1in,height=1.25in,clip,keepaspectratio]{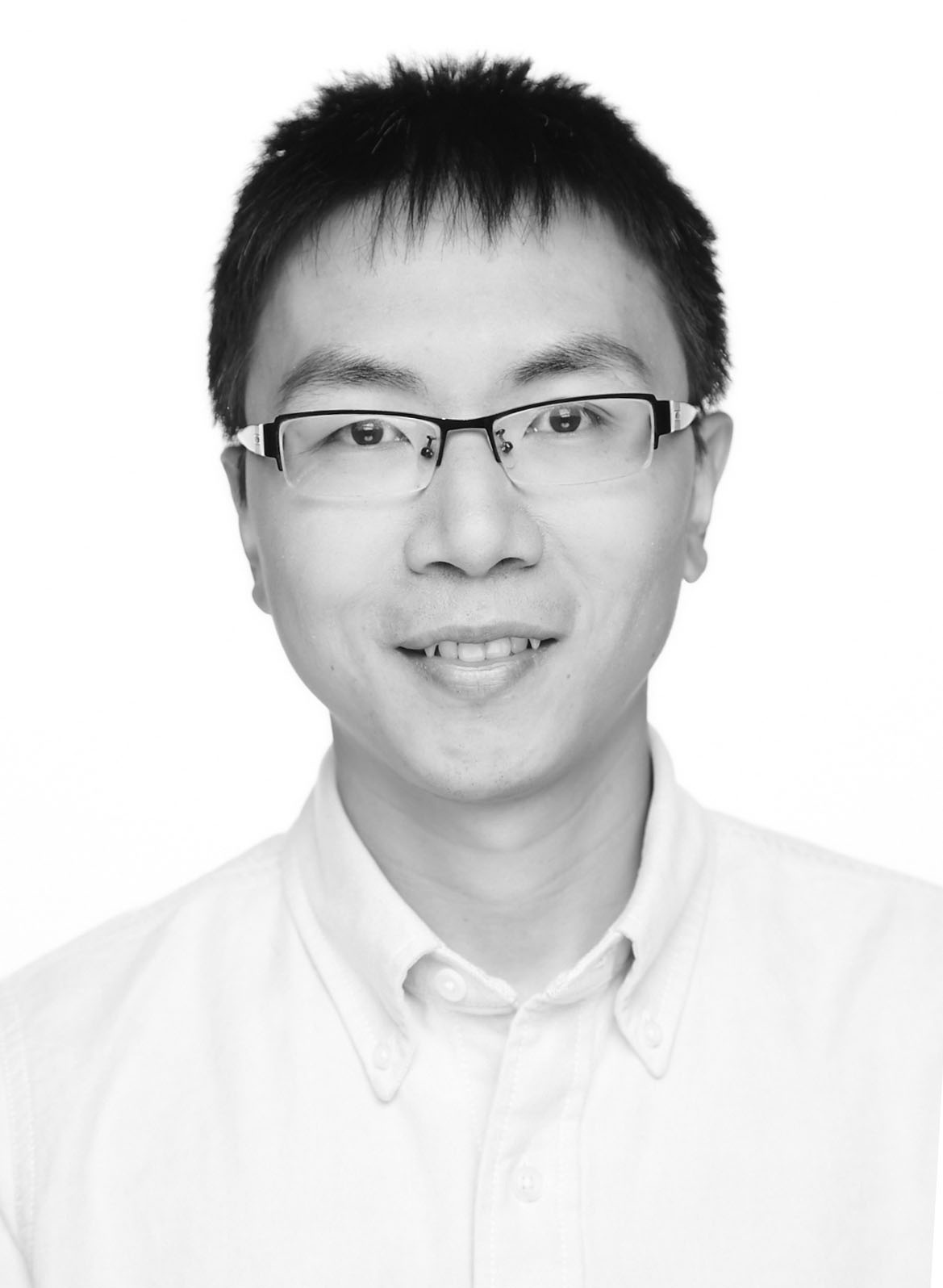}}]{Xiaohui Shen}
Xiaohui Shen is a Research Manager at ByteDance Inc.. Before that, he was a Senior Research Scientist at Adobe Research. He obtained his PhD degree from the Department of EECS at Northwestern University, and received the MS and BS degrees from the Department of Automation at Tsinghua University, China. His research interests include computer vision and deep learning.

\end{IEEEbiography}

\begin{IEEEbiography}[{\includegraphics[width=1in,height=1.25in,clip]{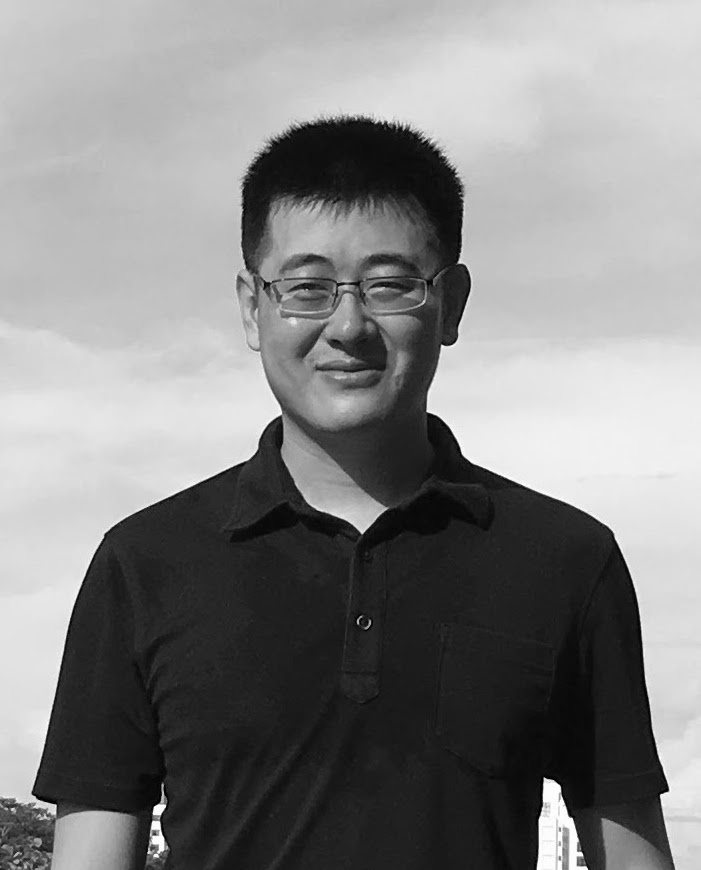}}]{Jiashi Feng}
is currently a research scientist at ByteDance. Before joining ByteDance, he was an assistant professor with the Department of Electrical and Computer Engineering at National University of Singapore and a postdoc researcher in the EECS department and ICSI at the University of California, Berkeley. He received his Ph.D. degree from NUS in 2014.  His research areas include deep learning and their applications in computer vision.  His recent research interest focuses on deep learning models, representation learning, and  3D vision. He received the best technical demo award from ACM MM 2012, best paper award from TASK-CV ICCV 2015, best student paper award from ACM MM 2018. He is also the recipient of Innovators Under 35 Asia, MIT Technology Review 2018. He served as the area chairs for NeurIPS, ICML, CVPR, ICLR, WACV, and program chair for ICMR 2017.

\end{IEEEbiography}





\end{document}